\documentclass[a4paper,fleqn]{cas-dc}

\usepackage[round]{natbib}
\usepackage{amsmath,amssymb,amsfonts}
\usepackage{siunitx}
\DeclareSIUnit\px{px}
\usepackage{subcaption}
\usepackage{adjustbox}
\usepackage{setspace}

\usepackage[colorinlistoftodos,prependcaption]{todonotes}
\usepackage{ulem} 
\usepackage{floatrow}
\usepackage{tikz}
\usetikzlibrary{tikzmark}

\usepackage{xcolor}
\usepackage{soul}

\begin{document}
\let\WriteBookmarks\relax
\def\floatpagepagefraction{1}
\def\textpagefraction{.001}
\shorttitle{3D shape sensing and deep learning-based segmentation of strawberries}
\shortauthors{Le Louëdec, Cielniak}

\title [mode = title]{3D shape sensing and deep learning-based segmentation of strawberries}

\author[1]{Justin Le Louëdec}[orcid=0000-0001-5387-9512   ]
\cormark[1]
\ead{jlelouedec@lincoln.ac.uk}

\author[1]{Grzegorz Cielniak}[
    orcid=0000-0002-6299-8465
]

\address[1]{Lincoln Centre for Autonomous Systems, University of Lincoln, Brayford Way, Brayford Pool, Lincoln LN6 7TS, United Kingdom}

\ead{gcielniak@lincoln.ac.uk}

\cortext[cor1]{Corresponding author}

\begin{abstract}
Automation and robotisation of the agricultural sector are seen as a viable solution to socio-economic challenges faced by this industry. This technology often relies on intelligent perception systems providing information about crops, plants and the entire environment. The challenges faced by traditional 2D vision systems can be addressed by modern 3D vision systems which enable straightforward localisation of objects, size and shape estimation, or handling of occlusions. So far, the use of 3D sensing was mainly limited to indoor or structured environments. In this paper, we evaluate modern sensing technologies including stereo and time-of-flight cameras for 3D perception of shape in agriculture and study their usability for segmenting out soft fruit from background based on their shape. To that end, we propose a novel 3D deep neural network which exploits the organised nature of information originating from the camera-based 3D sensors. We demonstrate the superior performance and efficiency of the proposed architecture compared to the state-of-the-art 3D networks. Through a simulated study, we also show the potential of the 3D sensing paradigm for object segmentation in agriculture and provide insights and analysis of what shape quality is needed and expected for further analysis of crops. The results of this work should encourage researchers and companies to develop more accurate and robust 3D sensing technologies to assure their wider adoption in practical agricultural applications.
\end{abstract}

\begin{keywords}
3D shape sensing \sep semantic segmentation for agriculture \sep machine learning architectures \sep simulation
\end{keywords}

\maketitle

\section{Introduction}

Modern agriculture is facing serious socio-economic challenges in its pursuit of sustainable food production. In particular, the shrinking labour force, caused by ageing populations, restrictions on migration or changing aspirations of agricultural workers, is causing major concern. Automation and robotisation of the sector are seen as a viable solution to this problem but requires several technical challenges to be solved to be successful~\cite{duckett18agricultural}.

One of the most important components of any automated agricultural technology is a reliable perception system providing information about crops, plants and the entire environment~\cite{mavridou2019machine}. For example, a vision system for detecting strawberry fruit can be used to monitor plant health, for yield forecasting but also for guiding a robotic arm picking individual fruit~\cite{from18rasberry}. Thanks to recent advances in 3D sensing technology and rapidly growing data-driven algorithms, the 3D vision has attracted considerable attention in recent years~\cite{dai2017scannet}. Compared to 2D images, 3D information provides additional depth cues critical for estimating the precise location and assessing the shape properties of various objects in the environment. So far, the main focus in the 3D vision community has been centred around benchmark datasets captured in controlled environments with large and rigid objects and fairly stable lighting conditions (e.g.~\cite{2017arXiv170201105A}). Although the preliminary deployment of 3D technology in agriculture has been reported~\cite{Vazquez_Arellano20163d}, only the sensing part was considered and therefore it is still unclear what are the main limitations of this technology operating in realistic scenarios. In particular, the efficiency of these sensing techniques to render shape information of the fruits and plants is of primary interest. The prior work on strawberries~\cite{boli,CLASSIFICATIONOFSTRAWBERRY,Nagamatsu}, considered mostly 2D images for analysis and focusing on the colour cue. Being able to obtain reliable 3D shape information from the field is a necessity for non-destructive and \textit{in situ} crop analysis undertaken by robotic technology developers, agronomists, biologists and breeders.

In this paper, we propose a study assessing the usefulness of the 3D shape information for horticultural produce growing in real farms, with semantic segmentation as a measurement of its importance and quality. The challenges posed by such a scenario include variable lighting conditions, reflections, occlusions, the non-rigid structure of the strawberry plants and the relatively small size of the fruit. Whilst the main focus of our work is on strawberry fruit, we also consider alternative crops in our analysis to highlight these challenges. Since the current 3D sensing technology has not been deployed widely in such scenarios and most of the modern machine learning algorithms were designed and trained specifically for large and rigid objects, our study aims to assess the usefulness and limitations of the sensing methods to render and utilise 3D shapes in agricultural scenarios. The paper also addresses the problem of an apparent lack of the existing public datasets with realistic 3D representations in agricultural domain which prevents fair comparisons and development of methods for this application area. In addition to experiments on real datasets, we propose the use of realistic simulation as a benchmarking tool for the proposed method similarly to~\cite{tian2018training}.

In particular, the contributions of this work are as follows:
\begin{itemize}
    \item Evaluation of the current 3D sensing technology, including time-of-flight and stereo cameras, for 3D shape sensing of soft fruit and comparison to other crops highlighting their unique shape properties;
    \item A novel deep network architecture designed for accurate and efficient semantic segmentation in 3D utilising as input points together with surface normal information;
    \item A realistic simulation of the soft fruit farm and 3D sensor modelling the surface reflectance of the objects, which is used as a reference point for evaluation of the existing 3D sensing technologies;
    \item Evaluation of the potential of the sensed 3D shape for recognising small objects such as soft fruit in cluttered outdoor scenes.
\end{itemize}

This paper builds on our initial results reported in~\cite{visapp20} which compared two competing 3D sensing technologies for the task of soft fruit detection. The current work introduces a thorough evaluation of the 3D sensing in real and controlled conditions, a scrutiny of the 3D shape sensing used for object segmentation in agriculture through the use of a simulated scenario and deployment of the superior machine learning architecture exploiting the organised nature of the 3D data originating from the evaluated sensors.

\section{Related work}\label{sec:related_work}

\subsection{3D sensing}\label{sec:3d_sensing}

Recent advances in 3D sensing technology and data-driven algorithms have resulted in a growing attention of the research community to the development of 3D vision systems~\cite{dai2017scannet}. In contrast to 2D images, 3D information provides directly the depth cues which are critical for estimating the precise location and assessing the shape properties of objects. Currently, the most popular 3D capturing devices are stereo cameras, time-of-flight (ToF) devices and LiDAR range finders (e.g. \cite{he2018advances,articletofstereo}). The stereo cameras use a pair of imaging sensors and calculate the correspondence between the two simultaneously captured images for estimating the depth. The stereo cameras based on visible spectrum require very stable lighting conditions and are therefore limited to selected indoor applications. To improve the robustness in less controlled conditions, the recent 3D stereo devices use infra-red sensors together with a projected pattern for improving disparity calculation in low-textured scenarios~\cite{test_cam}. On the other hand, the time-of-flight devices are based on emitting pulsating infra-red light for illuminating the scene and measuring the phase difference between the original and reflected pulses for estimating the distance. This sensing principle provides more accurate depth estimates when compared to the stereo sensors. Similar principle is used in LiDAR, which relies on a focused laser ray, resulting in the improved range and distance estimation compared to camera-based solutions. The low resolution or long acquisition time, however, are typically limiting the LiDAR's use for capturing and mapping large environments~\cite{wang2019survey} rendering them unsuitable for shape reconstruction of smaller objects such as agrcilutural crop.

Both stereo vision and ToF sensors operate in infra-red spectrum which in outdoor environments is highly affected by the natural variation of the infra-red light from the sun. In the agricultural domain, this issue is further amplified by low absorption and high reflectiveness of plant parts such as leaves or fruit~\cite{KAZMI2014128} together with the relatively small size of the considered objects~\cite{visapp20}. The issues with high reflectiveness tend to be more pronounced in stereo-based cameras affected by poor correspondence matching. One particular study highlighted this issue with strawberries which are particularly reflective in the infra-red spectrum~\cite{wlo,articlereflectance}.

\subsection{Deep learning for 3D information}

The core of machine learning methods using deep networks is applied to standard images, based on 2D convolutions which can be realised efficiently. The convolution operations in 3D require discretisation of space into so-called voxels, which is typically associated with some loss of information, large memory requirements and expensive computation rendering them unusable for most real-life scenarios. 3D processing using Deep Learning has been revolutionised by PointNet~\cite{qi2017PointNet} and subsequently by its improved variant PointNet++~\cite{qi2017pointnet2}. The PointNet architecture can be directly applied to a point cloud, through a segmentation/grouping of points in space using clustering algorithms. The basic PointNet architecture has been used to develop further improvements such as PointSIFT~\cite{jiang2018pointsift} or PointCNN~\cite{li2018pointcnn} leading to better discriminatory abilities but suffering from higher computational demands and not scaling well to real-time applications in realistic scenarios. The limitations of the PointNet architecture for agricultural scenarios which are characterised by noisy, cluttered and complex scenes were highlighted in our previous work~\cite{visapp20}. 

The information provided by the 3D camera-based sensors is organised into a regular grid which can be exploited for an efficient implementation of the convolutions. \cite{2016arXiv161108069L} chose to represent the information from the point cloud in a grid and encoded occupancy with a simple Boolean value. Such an approach allows for using standard 2D CNN architectures which demonstrated very good performance in object detection for autonomous driving. 
Other work~\cite{s19040893} makes the use of depth as a way to process spatial information. However depth does not encapsulate 3D shape information completely and by multiplying the number of neural networks by the number of inputs (3 inputs: RGB, depth and bird eye view representation), they reduce the efficiency and create scaling problems with higher resolutions of images.

In previous work, we have demonstrated the suitability of using the standard CNN architectures in 3D crop detection systems for robotic harvesters~\cite{Louedec_2020_CVPR_Workshops} using points and surface normals, reducing the network size and using directly shape and spatial information easily computed without learning. Such idea can be found in~\cite{rabbani2006segmentation}, where local surface normals and curvature combined with points connectivity allows segmentation of point cloud parts.

\subsection{3D vision and shape analysis in agriculture}

The majority of recent 3D vision algorithms and benchmarks were designed for indoor environments such as offices or residential rooms with large and rigid objects and fairly stable lighting conditions~\cite{dai2017scannet,2017arXiv170201105A}. When applied to an agricultural context in outdoor scenarios, however, these methods struggle to achieve satisfactory results~\cite{visapp20}, due to the noisy and sparse character of the 3D data originating from popular off-the-shelf RGBD sensors and issues with sensing in infra-read spectrum as described above. In agricultural applications, 3D information can provide important object characteristics including crop size, shape or location\\~\cite{Vazquez_Arellano20163d}. The most common approach to recognise such objects is based on a combination of 2D images for crop segmentation and detection and 3D information for augmenting the shape and location information. For example, \cite{2018arXiv181011920L} describes a perception system for harvesting sweet peppers. After scanning and reconstructing the scene using a robotic arm, the colour information together with point cloud is used to detect the pepper. The 3D projection of the segmented peduncle is then used to estimate a pose and the optimal grasping/cutting point. \cite{BARNEA201657} also presents a perception system for pepper harvesting but uses a colour agnostic method to detect fruit using depth information provided by RGBD cameras for localisation. By using highlights in the image, and 3D plane-reflective symmetries, the system is able to detect pepper fruit based on shape, in heavily occluded situations. \cite{Yoshida_2018jrm} use RGBD cameras and a two-level resolution voxelisation of the 3D space to detect tomato peduncles and the optimal cutting point for harvesting. The regions corresponding to tomatoes are first identified in the lower resolution, whilst dense voxelisation is used for the selected tomato regions to establish the optimal cutting points on the peduncle. A pure 3D crop detection system was proposed in \cite{Kusumam2017} for real-time detection of broccoli in the field which relied on hand-crafted features and SVM classifier. The suitability of using CNN architectures for the same application was later investigated in~\cite{Louedec_2020_CVPR_Workshops} with the superior results reported.

The shape of fruit or other crop is very important in plant breeding programmes as explained in~\cite{Nagamatsu} where phenotypic trait were identified with visual/genomic association. The shape information can be inferred from RGB images and/or depth images captured from RGBD cameras. In~\cite{LIN2021106107} the author propose to use RGB images for a prior segmentation before using the point clouds of the fruit and branches and fitting spheres and cylinders using the RANSAC algorithm to reconstruct the plant. \cite{SU2018261} presents a potato quality assessment system relying on 3D shape analysis, using a depth map captured in controlled light conditions. In~\cite{LI2017416} the focus is on assessing the shape of rice seed, with a high-precision laser scanner and the fusion of multiple point clouds. In more recent work~\cite{HAQUE2021106011}, an analysis of shape and quality for sweet potatoes is proposed, by capturing near infra-red and RGB images of the objects moving on a conveyor belt. The images are then used for 3D reconstruction and further shape analysis.A single fruit measurement system, by rotating a fruit on a platform to capture multiple point clouds using an RGBD camera and registering them into a single one for further morphological measurements is proposed in~\cite{agronomy10040455} . The shape identification and attributes of strawberries have been so far limited to colour 2D images~\cite{CLASSIFICATIONOFSTRAWBERRY}.

In field conditions, colour and 3D information can be use together to help improve the segmentation and recognition as in~\cite{WU2020105475} where an approach to combine colour and 3D geometry features is proposed to eliminate regions without fruit from point clouds to improve their localisation. In~\cite{LIU2020105621} we see a review of various hyper-spectral and 3D sensing methods to obtain precise plant phenotyping traits. it is however concluded that close-range hyper-spectral sensing techniques still needed lots of improvements due to poor precision and difficulties with plants such as non lambertians properties and self shadowed surfaces.

In contrast to the prior work, we present a thorough evaluation of the two most popular 3D sensing technologies applied to shape sensing of small agricultural objects such as soft fruit which pose particular challenges due to their size and high reflectance, together with a new approach to organised point cloud processing, using a CNN directly applied on the points coordinates and surface normals, placing a bigger emphasis on localisation, shape, and gemoetric structures. This approach addresses the problem encountered with un-organised approaches~\cite{qi2017pointnet2}, with faster inference times and improved feature extraction with facilitated point clustering thanks to the grid organisation of the features.

\section{Method}

Our methodology focuses on the study of shape information available through two different off-the-shelf 3D sensing technologies, and their usefulness for semantic segmentation of 3D point clouds in realistic agricultural scenarios. For the 3D segmentation task, we propose a novel deep network architecture which utilises the organised nature of information originating from the camera-based 3D sensors. The proposed segmentation algorithm is based on a classic auto-encoder architecture which uses 3D points together with surface normals and improved convolution operations. We propose using Transpose-convolutions, to improve localisation information of the features in the organised grid. We also present a set of baseline methods including a state-of-the-art 3D architecture~\cite{qi2017pointnet2} and a 2D-based method~\cite{segnet} used as a "measuring stick" and ultimate segmentation performance which should be aimed at with good 3D shape information. We also describe a realistic simulation of the soft fruit farm and 3D sensor modelling the surface reflectance of the objects, which is used as a reference point for evaluation of the shape information from real devices.

\subsection{3D sensing} \label{sec:sensors3d}
\begin{figure}
\centering
\begin{tabular}{c|c}
    \includegraphics[width=0.4\columnwidth]{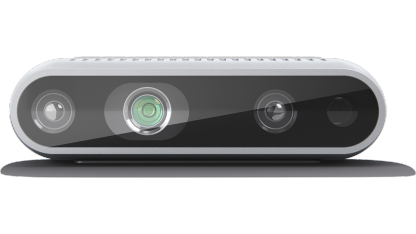}&
    \includegraphics[width=0.4\columnwidth]{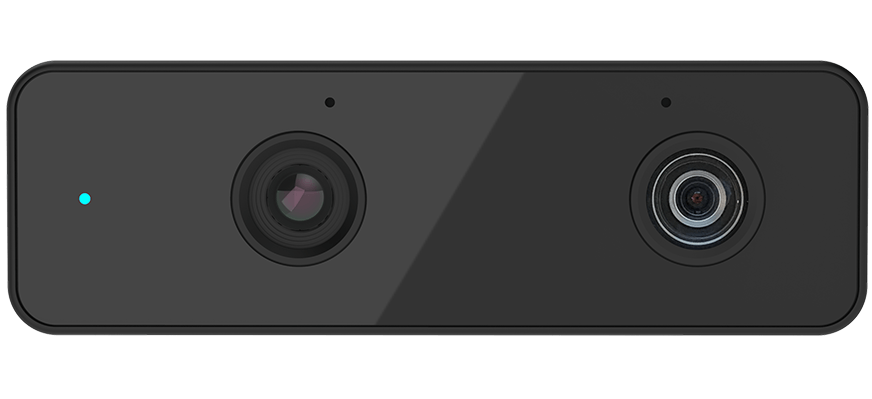}\\
    stereo & ToF\\
\end{tabular}
\caption{The selected 3D sensing devices: stereo camera Intel RealSense D435 (left) and ToF device PicoZense DCAM710 (right).} \label{fig:cameras}
\end{figure}

In this paper, we choose stereo and time-of-flight sensing technologies for sensing the shape of the strawberries in their natural growing conditions. The two selected cameras for evaluation are Intel RealSense D435 representing an infra-red stereo sensor and time-of-flight Pico Zense DCAM710 (see Fig.~\ref{fig:cameras}). Both cameras offer similar colour sensing capabilities with both featuring $\SI{1920}{\px} \times \SI{1080}{\px}$ RGB image resolution. The depth sensor of the RealSense device features $\SI{1280}{\px} \times \SI{720}{\px}$ resolution which is almost double compared to the Pico Zense camera ($\SI{640}{\px} \times \SI{480}{\px}$). The depth FOV for RealSense is $\ang{86} \times \ang{57}$ which covers slightly larger area than Pico Zense ($\ang{69} \times \ang{51}$). The RealSense device uses a stereo infra-red sensor pair together with the projected pattern for depth estimation allowing sensing in range of \SIrange{0.1}{10.0}{\metre}. The sensing range of the Pico Zense depth sensor is \SIrange{0.2}{5.0}{\metre}. In our application, the maximum sensing distance is restricted to \SI{0.8}{\metre} due to the small size of the strawberry fruit which is within the sensing range of both sensors.

All of these sensors provide depth information which can be de-projected into point clouds for a complete spatial information (x,y,z). We can however keep the organisation of points given by their depth map indices reducing the complexity of clustering the points.
These organised point clouds can be then used as inputs to our segmentation algorithms, and other easily computed (or captured) features can be added to the points coordinates, such as their corresponding surface normals or colour. We present our proposed segmentation method in the following section, which is using the data captured by such devices.

\subsection{CNN3D for organised point clouds}\label{sec:net}
\begin{figure}[ht]
    \centering
    \begin{tabular}{c|c}
        \includegraphics[trim=2cm 0 2cm 0, clip,width=0.45\columnwidth,height=0.25\columnwidth]{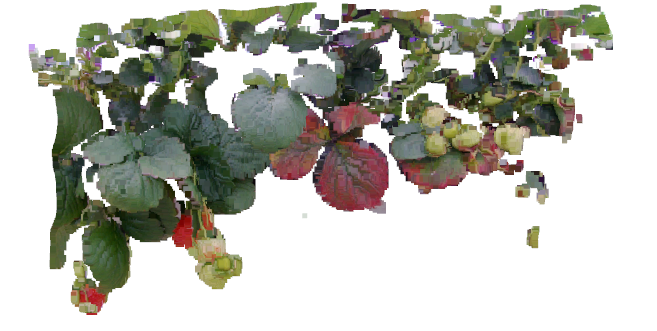}&
        \includegraphics[width=0.45\columnwidth,height=0.25\columnwidth]{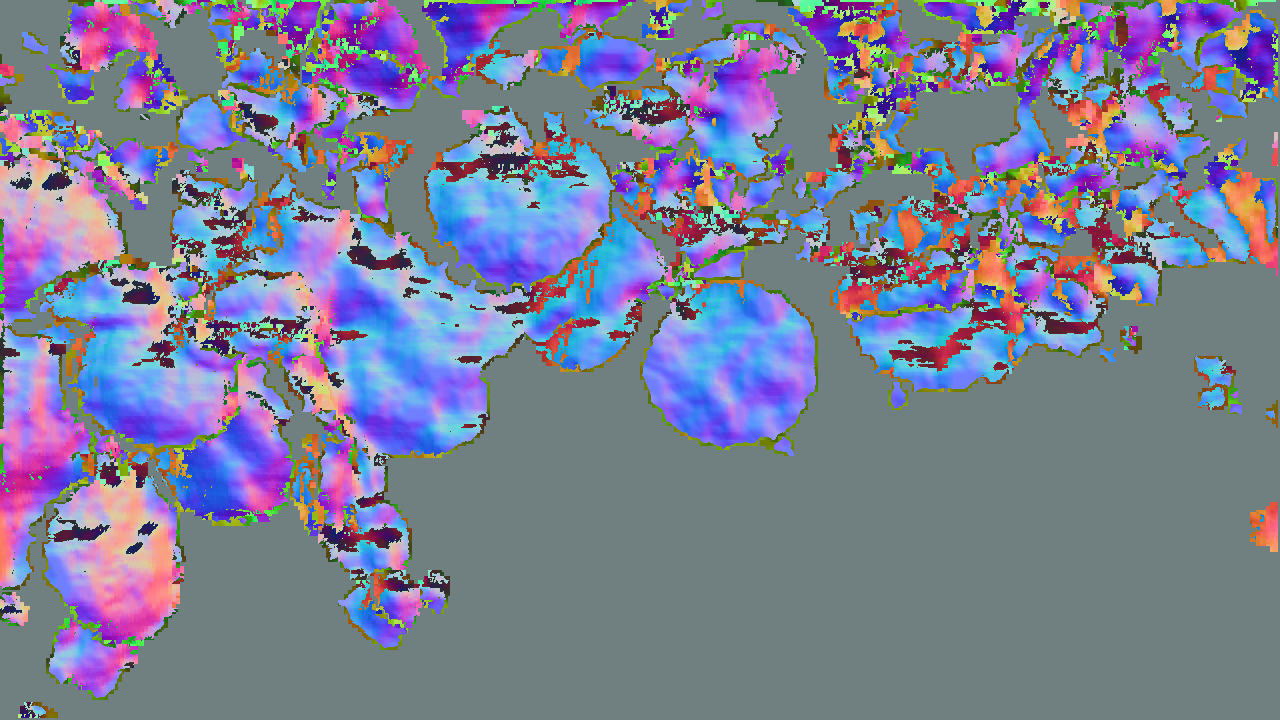}\\
    \end{tabular}
    \caption{A coloured point cloud from the ToF sensor (left column) and corresponding normal map (right column) used as an input for our CNN3D algorithm. The three normal components are mapped to the RGB format.}
    \label{fig:normal}
\end{figure}

For the semantic segmentation task, we chose a classic auto-encoder architecture inspired by U-Net~\cite{ronneberger2015u}, with the encoder part responsible for extracting relevant features and the decoder part for transforming them into the correct class prediction. We use skip connections between the encoding and decoding part of the architecture to make use of multi-scale features in the segmentation process \footnote{We provide the code for all the methods at : \url{https://github.com/lelouedec/PhD_3DPerception/}} (see Tab.~\ref{tab:network}). The input is composed of 6 features (\{X,Y,Z\} position of each 3D point and their 3-component normals \{\^{X},\^{Y},\^{Z}\} as presented in Fig.~\ref{fig:normal}), which are compressed into a $512 \times W \times H$ feature map in the latent space of the network, before being decoded into the segmentation mask. We use a standard VGG16 architecture, saving the feature maps and pooling indices at the end of four different convolution blocks. We use these indices in the decoder part of the network to up-sample the feature maps. Using un-MaxPool instead of traditional up-sampling, allows us to have a per point propagation of features, instead of interpolation in the grid. The saved feature maps are added to decoded feature maps in the decoder, to introduce multi-scale features and improve the extraction of point clusters corresponding to strawberries. The decoder blocs are inspired by the one proposed in SegNet~\cite{segnet}, where each convolutions are followed by BatchNormalisations and ReLu. Due to the compact nature of the CNNs, the inference time exceeds real time. A similar architecture was used in our previous work for the problem of 3D broccoli detection~\cite{Louedec_2020_CVPR_Workshops}. This architecture differs from other 3D segmentation techniques by using 2D convolutions directly over organised 3D information and the use of multi-scale feature maps combination approach. Using an organised version of the point cloud, makes the clustering of points faster and more efficient than methods such as PointNet. Indeed points from the same object will be in the same region of the grid, which is faster and easier to run through than finding points directly in space. We use feature maps at different stages of their encoding (different size and scale of features), in the decoding part of the architecture. This allows us to take in consideration different size and scale of features and objects to separate strawberries from the background. We can also add colour as an extra feature to the input resulting in 9 dimensions \{X,Y,Z,\^{X},\^{Y},\^{Z},R,G,B\} (points coordinates, normal and colour). We use colours as a third feature input as a mean to compare the importance of shape and colour information for the segmentation of strawberries.
For training we use as a loss function a weighted Cross entropy loss, with the Adam optimiser~\cite{kingma2014adam}. We use the percentage of points belonging to strawberries in each point clouds to compute the weight for each classes used with Cross entropy loss.
We use the same training strategy to train the baselines architectures.

\newcolumntype{C}[1]{>{\centering\arraybackslash}p{#1}}
\begin{table*}[ht]
  \centering
    \begin{tabular}{cc}
        \begin{tabular}{|C{6cm}|}
            \hline
            \tikzmark{s}Encoder\\
            \hline
            Input : Points coord. + Normals \\ \hline
            6-conv3-64\\
            64-conv3-64\\
            \hline
            MaxPool-2-2\tikzmark{g}\\
            \hline
            64-conv3-128\\
            128-conv3-128\\
            \hline
            MaxPool-2-2\tikzmark{e}\\
            \hline
            128-conv3-256\\
            256-conv3-256\\
            256-conv3-256\\
            \hline
            MaxPool-2-2\tikzmark{c}\\
            \hline
            512-conv3-512\\
            512-conv3-512\\
            512-conv3-512\\
            \hline
            MaxPool-2-2\tikzmark{a}\\
            \hline
            512-conv3-512\\
            512-conv3-512\\
            \tikzmark{s2}512-conv3-512\\
            \hline
        \end{tabular}
 & 
        \begin{tabular}{|C{6cm}|}
            \hline
            Decoder\\
             \hline
            Output: Point classes pred.\tikzmark{z2}\\
            \hline
            64-Tconv3-2\\
            64-Tconv3-64\\
            \hline
            \tikzmark{h}UnMaxPool-2-2\\
            \hline
            128-Tconv3-64\\
            128-Tconv3-128\\
            \hline
            \tikzmark{f}UnMaxPool-2-2\\
            \hline
            256-Tconv3-128\\
            256-Tconv3-256\\
            256-Tconv3-256\\
            \hline
            \tikzmark{d}UnMaxPool-2-2\\
            \hline
            512-Tconv3-256\\
            512-Tconv3-512\\
            512-Tconv3-512\\
            \hline
            \tikzmark{b}UnMaxPool-2-2\\
            \hline
            512-Tconv3-512\\
            512-Tconv3-512\\
            512-Tconv3-512\tikzmark{z}\\
            \hline
        \end{tabular}
 \\
 \tikzmark{s3} & \tikzmark{s3b}\\
    \end{tabular}
    \caption{The CNN3D architecture used for semantic segmentation. For standard convolutions and transpose convolutions the format is: input size $-$ kernel size $-$ output size. For maxpooling and un-maxpooling: kernel size $-$ stride. The arrows indicate skip layer connections between encoder and decoder. }\label{tab:Network}
    \begin{tikzpicture}[overlay,remember picture]
\draw[->,black,-latex,very thick] ([shift={(0.3,0.1)}]pic cs:a) to ([shift={(-0.1,0.1)}]pic cs:b) ;
\draw[->,black,-latex,very thick] ([shift={(0.3,0.1)}]pic cs:c) to ([shift={(-0.1,0.1)}]pic cs:d) ;
\draw[->,black,-latex,very thick] ([shift={(0.3,0.1)}]pic cs:e) to ([shift={(-0.1,0.1)}]pic cs:f) ;
\draw[->,black,-latex,very thick] ([shift={(0.3,0.1)}]pic cs:g) to ([shift={(-0.1,0.1)}]pic cs:h) ;
\draw[->,black,-latex,very thick] ([shift={(-2.2,-0.2)}]pic cs:s) to ([shift={(-1.7,0.0)}]pic cs:s2) ;
\draw[->,black,-latex,very thick,bend right=10] ([shift={(0.0,0.0)}]pic cs:s3) to ([shift={(0.0,0.0)}]pic cs:s3b) ;
\draw[->,black,-latex,very thick] ([shift={(1.5,0.0)}]pic cs:z) to ([shift={(0.8,0.25)}]pic cs:z2) ;

\end{tikzpicture}
\end{table*}

\subsection{Baseline architectures}

We use a modified PointNet++ architecture~\cite{qi2017pointnet2} as the 3D baseline for comparisons to CNN3D. The architecture can process point clouds directly by grouping points using algorithms such as K-Nearest Neighbours (KNN) or ball query before applying simple convolutions on  features from these clusters. PointNet++ can be separated into two main layers. First the set Abstraction layer (SA), extracts features from the point cloud by considering for each point their neighbourhood using a predefined radius. Secondly the Feature Propagation (FP) layer, interpolates features and learns decoding into the dimension of the targeted SA layer all the way up to the same size as the input point cloud. For a segmentation task, the latent space is fed into a succession of FP layers up to the original point cloud size, to decode for each point the class it belongs to. This can be seen as a classical encoder/decoder architecture with the SA layer succession being the encoding part and the FP layers succession the decoding part. Colour is added as a feature which is concatenated to points spatial information in  the first SA layers. We provide the architecture summary for PointNet++ in Table~\ref{tab:network} with Abstraction layer (SA) and Feature Propagation (FP) layers details.

\begin{table}[ht]
\caption{The PointNet++ configuration.}\label{tab:network} \centering
\begin{tabular}{|c|c|c|c|}
 \hline
 \multicolumn{4}{|c|}{encoder}\\
 \hline
 layer & \#points & radius & mlps\\
 \hline
 SA & 4096 & 0.1 & [16, 16, 32]\\
 SA & 2048 & 0.1 & [32,32,32]\\
 SA & 1024 & 0.1 & [32,32,64]\\
 SA & 256 & 0.2 & [64,64,128]\\
 SA & 64 & 0.4 & [128,128,256]\\
 SA & 16 & 0.8 & [256,256,512]\\
 \hline
\end{tabular}
\quad
\begin{tabular}{|c|c|}
 \hline
 \multicolumn{2}{|c|}{decoder}\\
 \hline
 layer & features\\
 \hline
 FP & 256,256\\
 FP & 256,256\\
 FP & 256,128\\
 FP & 128,128,128\\
 FP & 128,128,64\\
 FP & 128,128,64\\
 MLP & [64,128]\\
 MLP & [128,2]\\
 \hline
\end{tabular}
\end{table}

Since the majority of vision systems for agriculture employ standard colour cameras, we also propose to compare our 3D system to a 2D baseline. As a potential limit for segmentation algorithms we use the superior results achieved through the use of a classic SegNet architecture~\cite{segnet} which is based on an encoding/decoding principle. The encoding part creates latent feature maps extracted from the input whilst the decoder transforms the features into a segmentation mask, attributing for each pixel its predicted class. Our implementation uses batch normalisation~\cite{10.5555/3045118.3045167}, ReLU as an activation function and max pooling for dimensionality reduction and compression into the latent space. In the decoding part of the architecture, up-sampling is done using an un-max pooling function with the indices taken from the encoding part of the architecture. Finally a softmax function is used to transform the output of the network to a per-class probability for each pixel.

\subsection{Simulated benchmarking environment}\label{sec:simu}
\begin{figure}[ht]
\centerline{\includegraphics[trim=150pt 20pt 150pt 20pt, clip, width=\columnwidth]{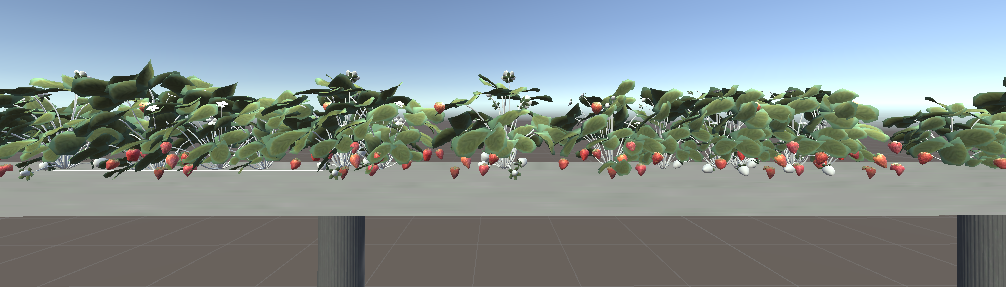}}
\caption{The realistic simulation of a table top strawberry farm implemented in the Unity engine.} \label{fig:sim}
\end{figure}

To fully assess the limitations of the current 3D sensing technology and the proposed segmentation method, we propose to introduce a realistic simulated environment featuring a 3D sensor with characteristics and parameters similar to the real devices but allowing for control of the depth sensing quality. The inherent part of the simulation is a realistic environment, which in our case is based on a real strawberry farm (see Fig.~\ref{fig:sim}) and implemented in the Unity game engine~\cite{engine2008unity}. Unity offers an integrated physics and 3D graphics simulator allowing for realistic renderings of natural scenes, different light conditions, shadows, textures together with a number of tools streamlining the development and data annotation process.
The simulated device implements a colour (i.e. RGB) and depth sensor with the same resolution of $\SI{871}{\px} \times \SI{530}{\px}$ and a field of view of $\ang{60}$. The selected parameters result in a number of 3D points similar to both real sensors. The depth sensing range was set to \SIrange{0.3}{2.0}{\metre}.

Simulating physically accurate depth sensors requires prior manual measurements of the objects of interest as illustrated in Bulczak et al.~\cite{s18010013}. The presented approach provides a physically accurate simulation of a ToF camera, mainly through the measurement of the Bidirectional Reference Distribution Function (BRDF), describing how light is reflected at this opaque surface, in this case near-infra red light, which is mostly used for ToF cameras. However, BRDF data for different materials and objects is acquired through a very complex process explained in Dana et al.~\cite{10.1145/300776.300778}. Due to the lack of precise data available for strawberries and plants, we focus on approximating the behaviour of reflectance through the normal information and insight from Liu et al.~\cite{reflectance}. The reported reflectance between 70\% and 80\% in the infrared wavelength ($[750,900]nm$) used by most ToF cameras. Furthermore, Kolb et al.~\cite{tofgraphics} describe over-reflective material to be leading to saturation and low reflective material to produce low signal. The influence of the reflectance on the effectiveness of LiDAR sensors is evaluated in simulated environments by Muckenhuber et al.~\cite{s20113309}.

\begin{figure}
\centerline{    \includegraphics[width=0.7\linewidth]{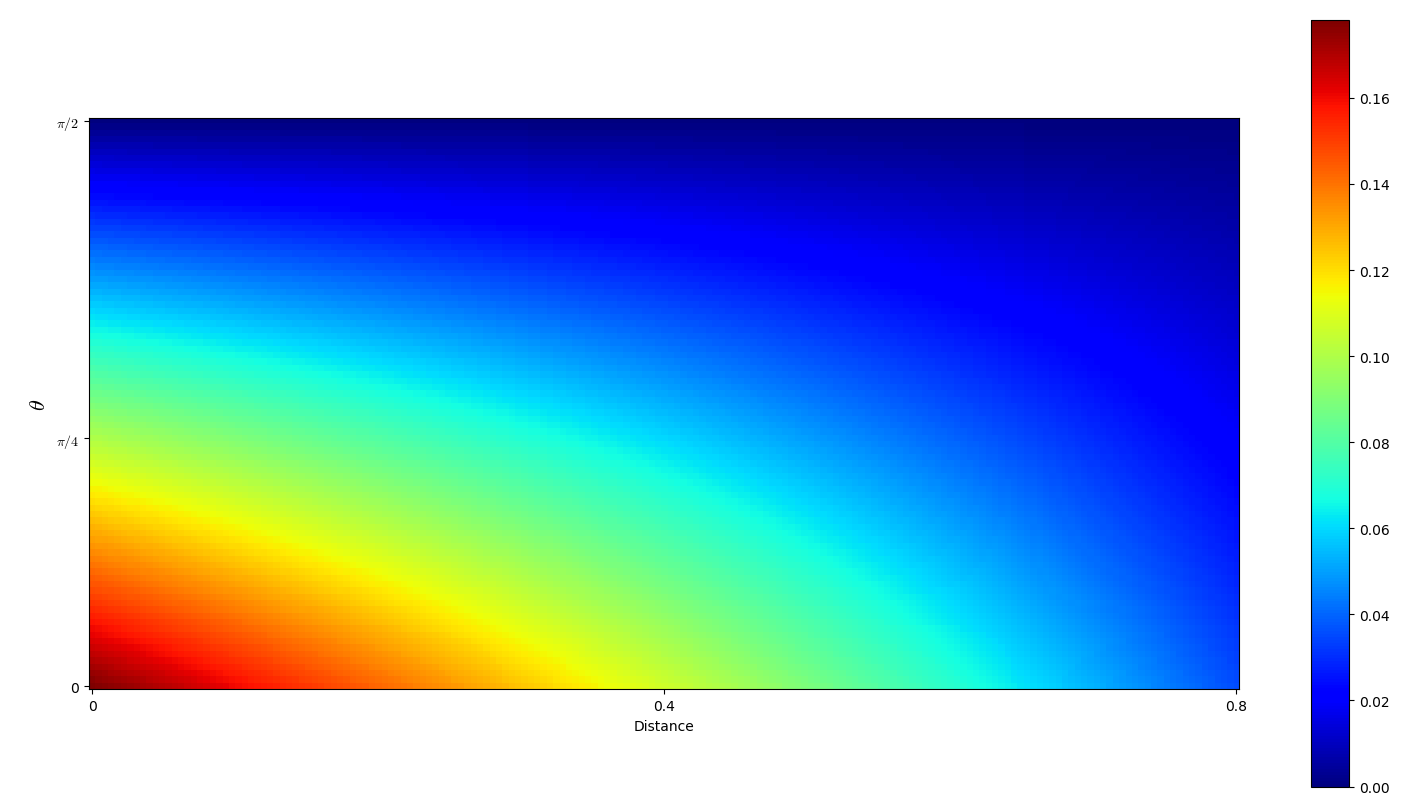}}
\caption{The influence of the angle of incidence $\theta$ and distance from the camera on the surface reflectance values.} \label{fig:plotreflec}
\end{figure}

We model the reflectance of the soft fruit based on the information from~\cite{reflectance,tofgraphics}, and add surface distortion based on surface normal and incidence direction of the camera's emitted light. The disturbance $\Delta P$ is expressed as
 \begin{equation}
     \Delta P = \Delta v \left( 1 - \frac{\theta + \frac{\pi}{2}}{\pi} \right)
     \label{eq:distortion}
 \end{equation}
where $\Delta P$ is the amount of change over the point position, $\Delta v$ a ratio to the distance from the camera (points further away receive less distortion as the saturation is lower), and $\theta$ is the angle between the surface normal at this point and the ray of light from the camera. We normalise this angle between 0 and 1, to have the distortion dependant on the direction faced by the surface (surface facing directly the camera are more prone to higher saturation) with higher changes for angles close to 0 (parallel to light direction), and lower for angles close to $\frac{\pi}{2}$. We illustrate this reflectance in Fig.~\ref{fig:plotreflec}

\begin{figure}
\centering
\begin{tabular}{c|c}
    \includegraphics[width=0.35\columnwidth,height=0.35\columnwidth]{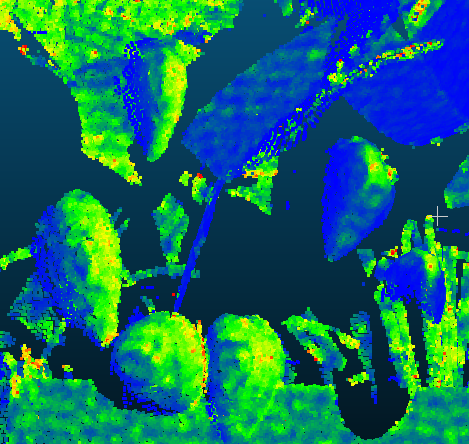}&
    \includegraphics[trim=3cm 2.5cm 1.8cm 2cm, clip,width=0.35\columnwidth,height=0.35\columnwidth]{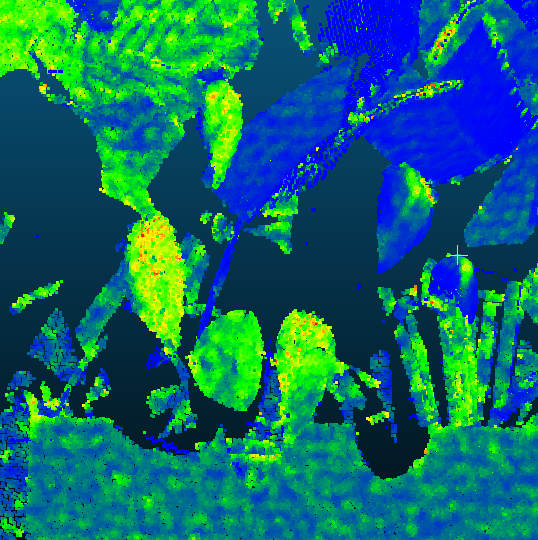}\\
\end{tabular}
\caption{Effect of the reflectance added to the simulation: the normal change rate with zero mean Gaussian noise (left), and the same scene with the reflectance added (right).} \label{fig:reflectance}
\end{figure}

 We also use an additive zero mean Gaussian noise $\mathcal{N}(0,\sigma^{2})$ to simulate imperfections in depth reconstruction. This noise is meant to simulate electronics and sensor noise, with a global illumination. It smooths out the very linear depth captured, adding randomness over surfaces and rendering it less perfect but keeps shapes close to the original models.
 We show the difference made by the reflectance over the noise in Fig.~\ref{fig:reflectance}.
 
\begin{figure}
\centering
\begin{tabular}{c|c|c|c}
    \includegraphics[width=0.22\columnwidth,height=0.2\columnwidth]{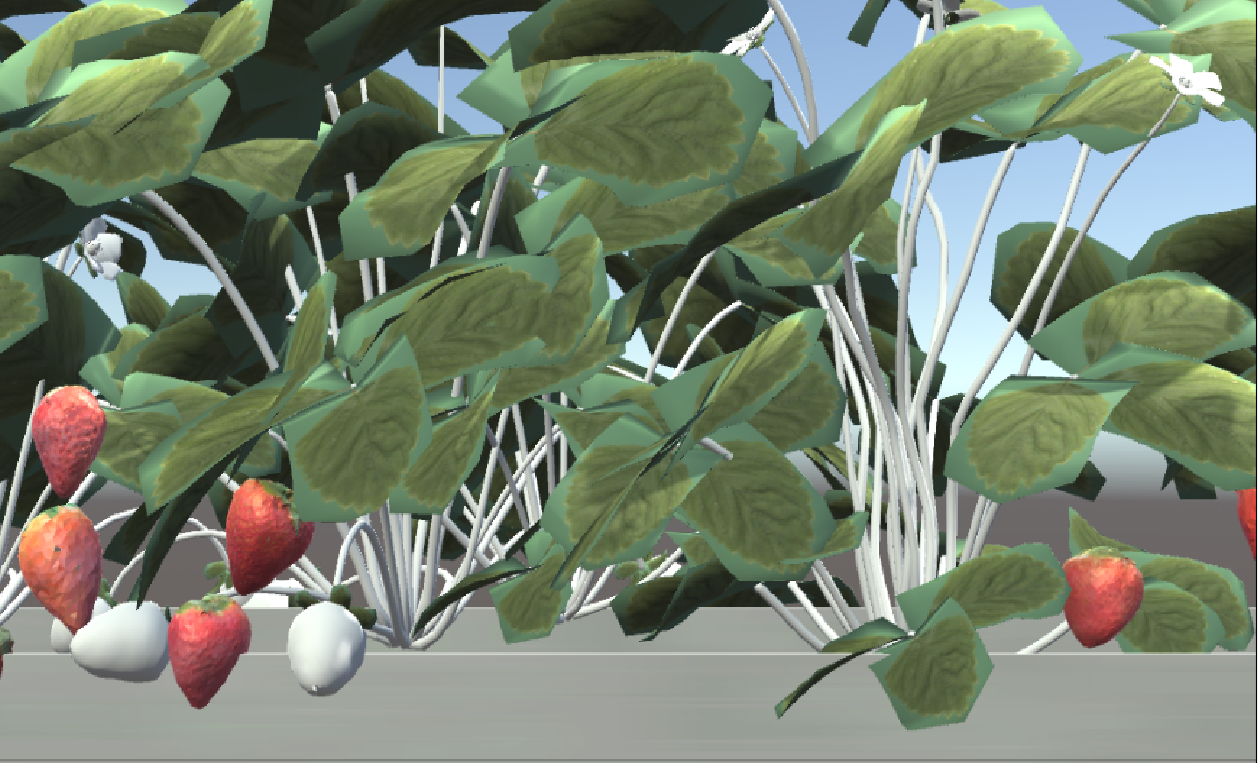}&
    \includegraphics[width=0.22\columnwidth,height=0.2\columnwidth]{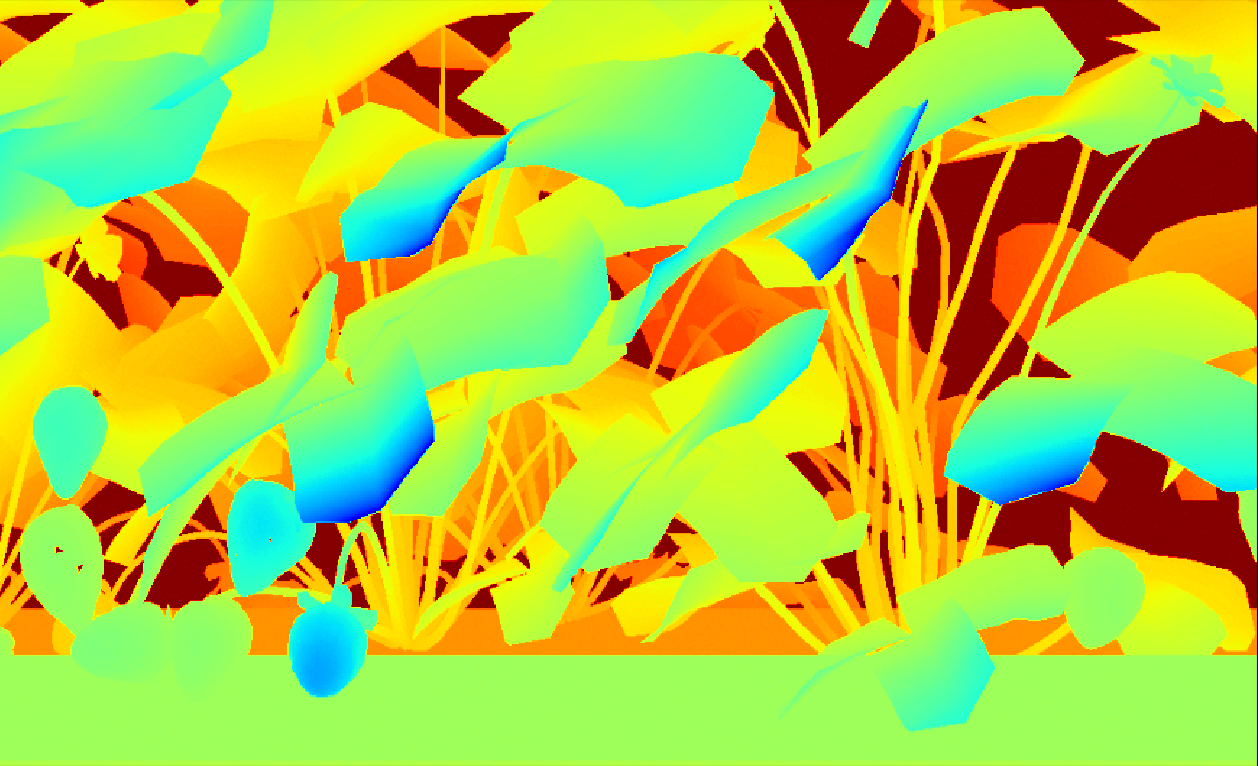}&
    \includegraphics[width=0.22\columnwidth,height=0.2\columnwidth]{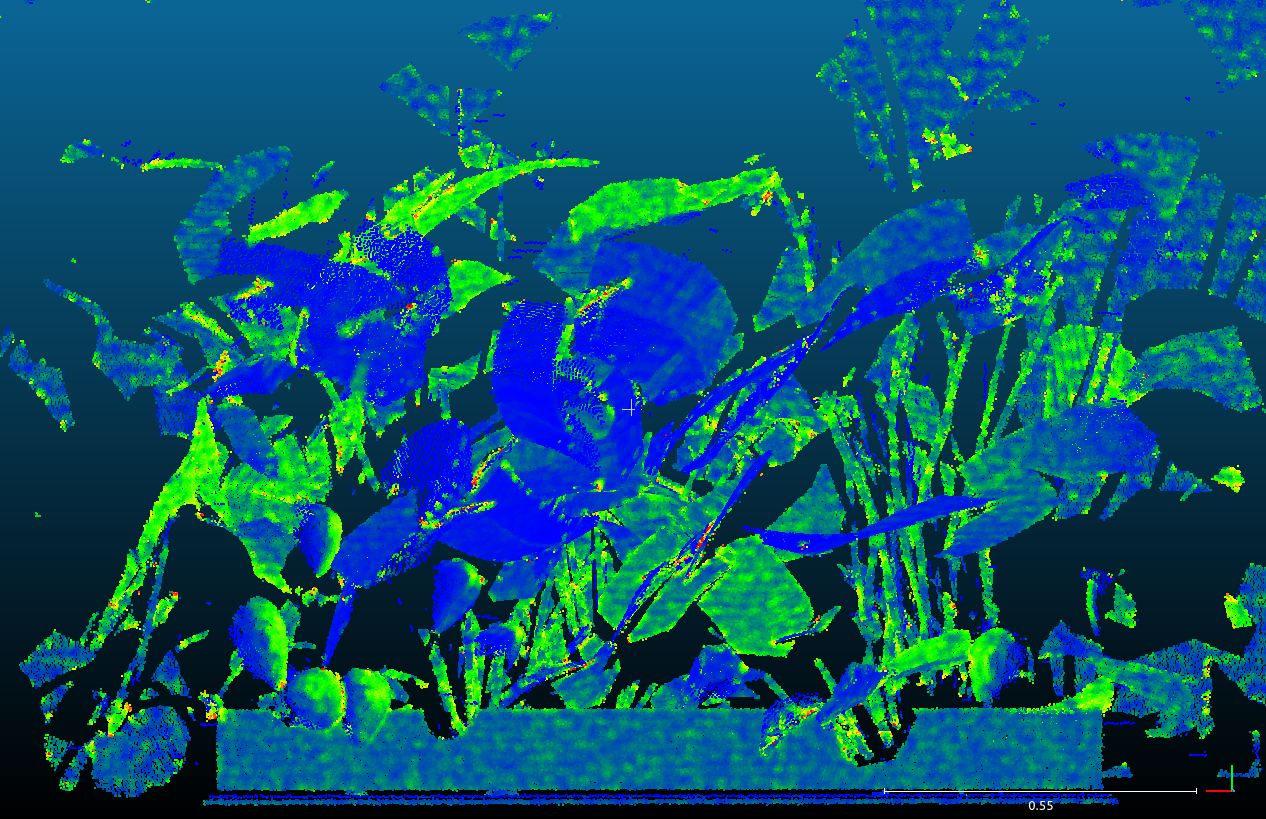}&
    \includegraphics[width=0.22\columnwidth,height=0.2\columnwidth]{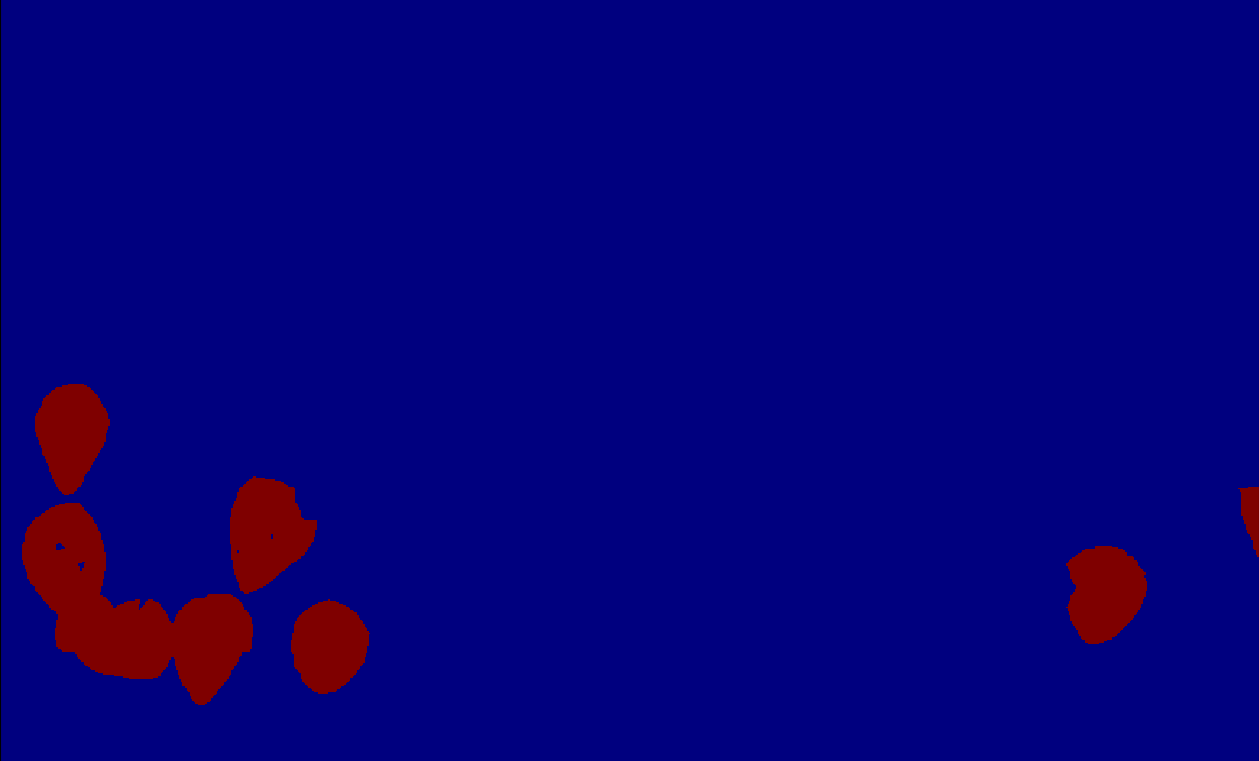}\\
    a & b & c & d\\
\end{tabular}
\caption{A side view of the strawberry farm as seen through the simulated sensor including colour image (a), depth image (b), curvature map (c) and ground truth annotation (d).} \label{fig:unitydata}
\end{figure}

Our simulation environment features a mock-up strawberry farm consisting of a single $35$ m long row of table tops and strawberry plants with flowers and berries attached to the stems using 9 different commercially available 3D models augmented by realistic scans of real strawberries from~\cite{boli}. The individual plants are multiplied and randomly positioned along the tabletop row creating realistic overlaps between leaves, stems, plants and berries. The strawberries are also grouped into clusters to simulate the appearance of real varieties used in our experiments.

A snapshot of an example scene from the simulated farm as perceived by the sensor, together with an automated semantic segmentation annotation can be seen in Fig.~\ref{fig:unitydata}.

\section{Evaluation Methodology}
We evaluate shape and depth produced by the previously presented sensors, through the impact on semantic segmentation as well as a quantitative and qualitative study of fruit shape. We first present in Sec.~\ref{sec:datasets} the different datasets used for these various analysis and in Sec.~\ref{sec:Metodores} the different metrics used to evaluate semantic segmentation. Finally we present various shape quality indicators for the berries in Sec.~\ref{sec:shapeeval}, from roundness and average ellipsoid shape to the use of spherical harmonics coefficients. We also introduce the normal surface change indicator, used in Sec.~\ref{sec:qualishape}, to offer a visual analysis of the shape captured in different datasets.

\subsection{Datasets}\label{sec:datasets}

\begin{figure}
\vspace{0.1cm}
\centering
  \begin{subfigure}[b]{0.45\columnwidth}
    \includegraphics[trim=20cm 10cm 20cm 45cm, clip, width=\columnwidth]{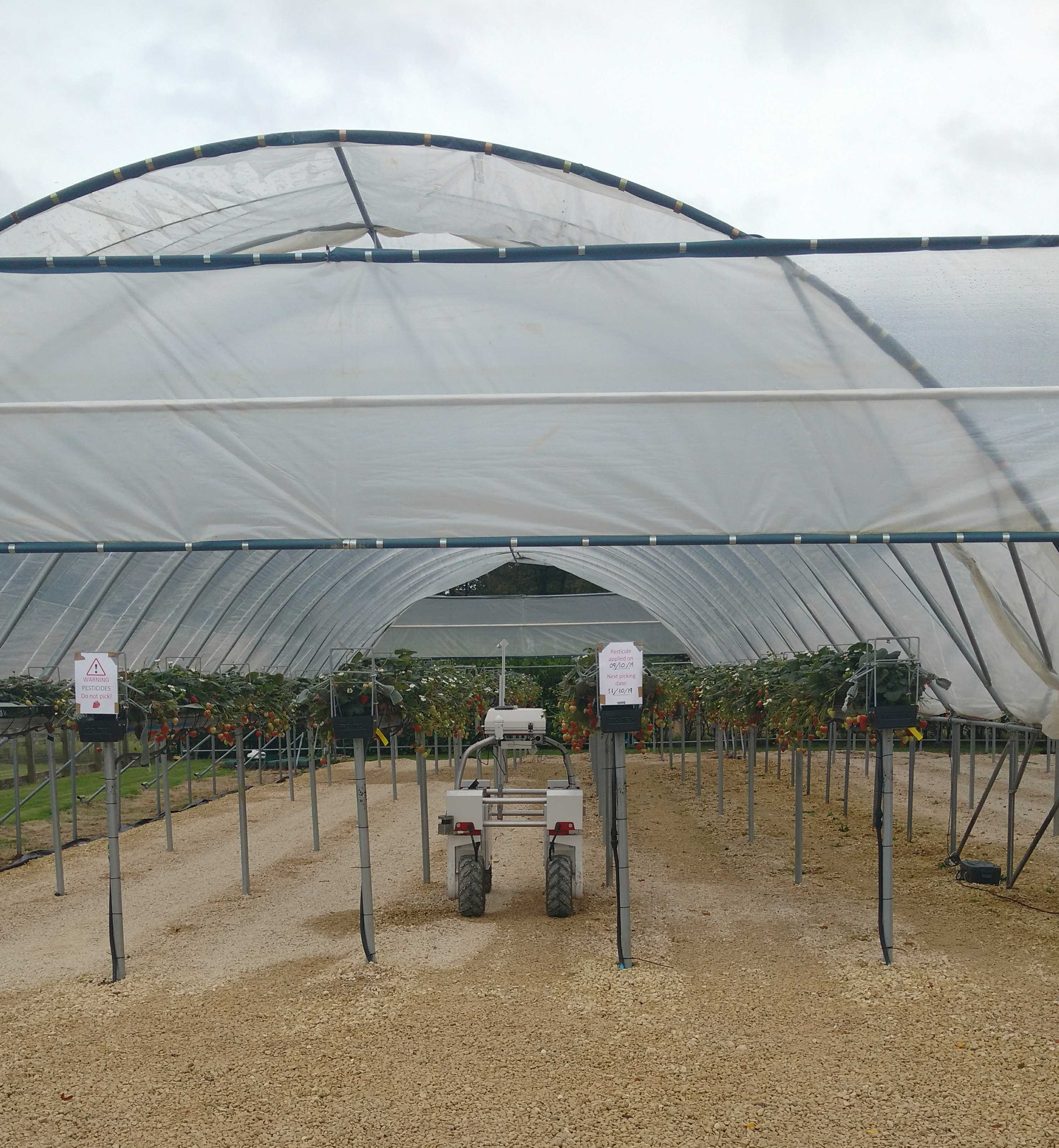}
    \label{fig:1}
  \end{subfigure}
  \begin{subfigure}[b]{0.45\columnwidth}
    \includegraphics[width=\columnwidth]{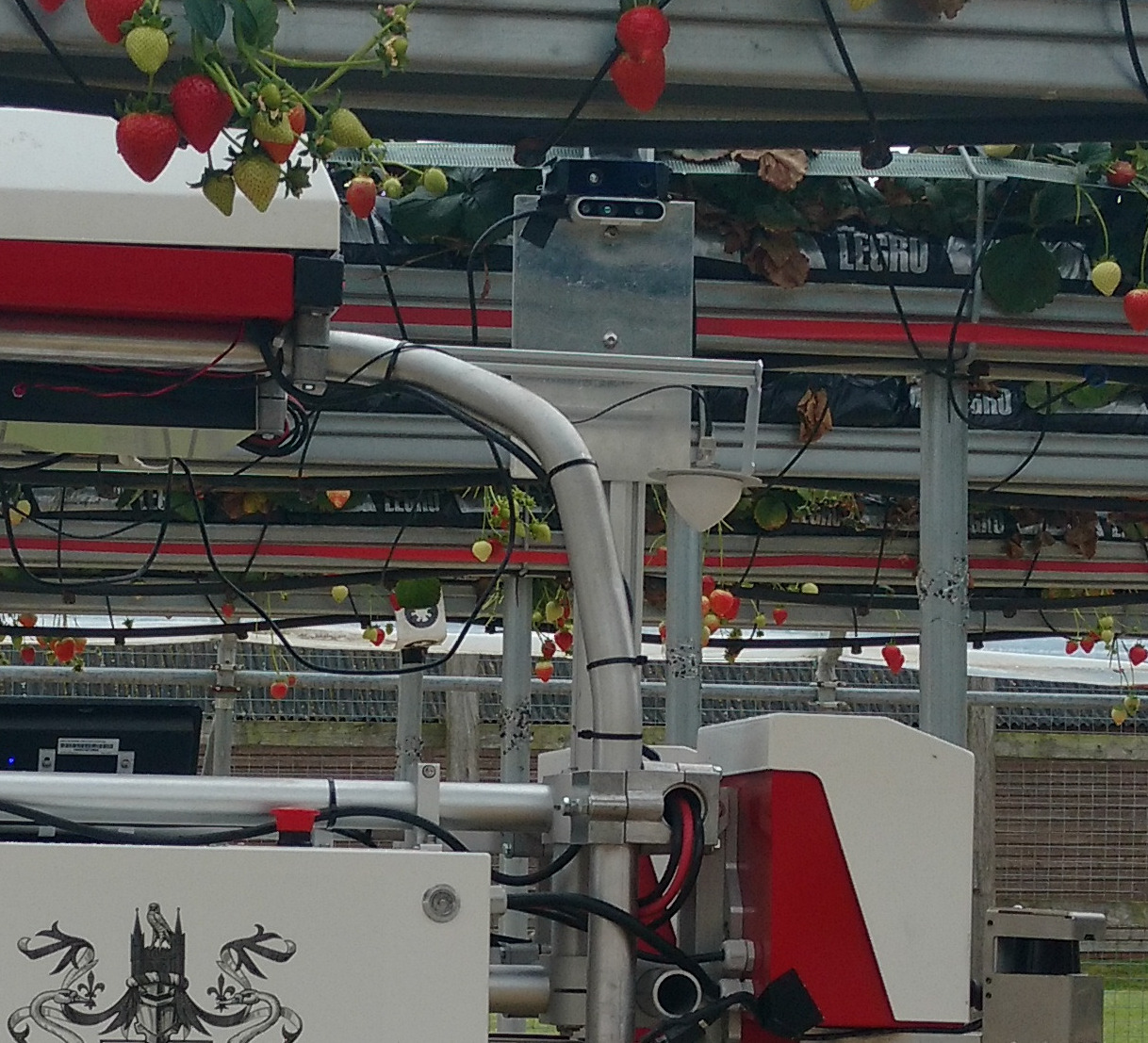}
    \label{fig:2}
  \end{subfigure}
  \caption{The strawberry farm, with a robot roaming in the tabletop rows collecting data (left). The sensor set-up used for data collection (right).}\label{fig:polytu}
  \vspace{-0.1cm}
\end{figure}

To evaluate the selected 3D sensing devices for the task of segmentation of strawberry fruit, we used them to collect datasets from a real environment. To that end, we have deployed our data acquisition system at a mini version of a real strawberry farm, located at the Riseholme campus of the University of Lincoln. The farm features two polytunnels of 6 tabletop rows, \SI{24}{\meter} long with an industrial variety of strawberries (everbearer Driscoll’s Amesti) as depicted in Fig.~\ref{fig:polytu}.  

The data capture setup, featuring the stereo and ToF sensors, was mounted on an agricultural robot Thorvald~\cite{grmistad17thorvald}. The robot autonomously navigated the polytunnel rows, stopping every 20 cm to collect a snapshot from both views (see Fig.~\ref{fig:polytu}). The capturing session took place in October 2019 and resulted in colour images and point clouds representing different growth stages of plants and fruit. The datasets were then manually annotated to indicate the location of strawberry fruits resulting in 132 labelled point clouds with around 1574 instances of ripe strawberries for ToF data and 57 point clouds for around 750 instances for stereo data (see Table~\ref{tab:datares}). The simulated dataset was created to mimic the captured data described above. We automatically captured a number of point clouds with the attributes and parameters similar to the real cameras used in the field. The noise parameter was experimentally tested for different levels of $\sigma$ in range of $[0.5-1.0]$ and due to its negligible influence on the segmentation performance set to $\sigma=\SI{0.5}{\centi\meter}$. The reflectiveness value was set to $\Delta v=0.2$ for a disturbance between 0.7 cm for the closest and perpendicular surfaces and 0.0 cm for the farthest and perpendicular surfaces.
To assure a comparable size of the datasets, we balanced out the total number of points for each sensor. The stereo and ToF dataset were manually annotated by creating pixel-level annotation in RGB images which were then projected into the aligned point clouds. Annotation of the simulated data was also performed at per pixel-level, but thanks to its nature, was performed fully automatically. The summary of the datasets collected for the experiments is presented in Table~\ref{tab:datares}.

\begin{table}[ht!]
\caption{The summary of datasets collected.}\label{tab:datares} \centering
\begin{tabular}{|c|c|c|c|}
  \hline
  sensor & stereo & ToF & simu \\
 \hline
 \# point clouds & 57 & 132 & 134\\
 \# points & $\sim$ 300k & $\sim$ 300k  & $\sim$ 400k\\
 range & 20cm-65m & 20cm-70cm & 20cm-80cm\\
 \# instances & $\sim$750 & $\sim$1574 & $\sim$1200\\
 \% straw points & 4.9\% & 3.0\% & 4.5\% \\
 \hline
\end{tabular}
\end{table}

For shape evaluation, we select a subset of 10 strawberry samples from~\cite{boli} which were used to create the realistic simulated fruit models described in Sec.~\ref{sec:simu}.
Finally we use two examples from the fuji dataset~\cite{GENEMOLA2020105591} and broccoli dataset~\cite{kusuman,Louedec_2020_CVPR_Workshops}. These datasets offer different capturing techniques, and focus on larger fruits/vegetables. They allow us for a comparison of potential goals for shape capture quality in Sec.~\ref{sec:qualishape}.

For all the dataset we use 80\% for training and the remaining 20\% for testing for the segmentation task. For the simulated data, we also use data augmentation, by randomly rotating the point clouds around the z-axis, between -180 and 180 degrees.

\subsection{Evaluation metrics for segmentation}\label{sec:Metodores}

To evaluate our trained models, we use standard semantic segmentation metrics including Accuracy, mean Intersection over Union (mean IoU) and Cohen's Kappa~\cite{doi:10.1177/001316446002000104}. The Accuracy $Acc = \frac{TP+TN}{P + N}$ measures how accurate the prediction is compared to the ground truth, without taking into account the class imbalance. The mean $IoU = \frac{TP}{(TP+FP+FN)}$ is the overlap of the output predicted by the algorithm with the ground truth and averaged for every class and samples. The Cohen's Kappa coefficient is particularly useful for unbalanced data, where one class is more represented than the others - in our case, background represents the majority of points when compared to strawberries. This measure provides a better assessment of the real discriminatory power of the classifier and takes the observed and expected accuracies into account: $\kappa= \frac{(Acc_{obs} - Acc_{exp})}{1 - Acc_{exp}}$. The Observed Accuracy $Acc_{obs}$ is the number of instances correctly classified, and the Expected Accuracy $Acc_{exp}$ is what any random classifier should be expected to achieve. 
We finally use the area under the precision-recall curve (AUC) as additional metric for comparisons between different methods.

\subsection{Shape quality indicators}\label{sec:shapeeval}

To evaluate the quality of shape information in our scenario, we propose to use several metrics which are adequate for the shape of the strawberry. We use a roundness metric from~\cite{CRUZMATIAS201928}, which is expressed as the deviation from the minimum volume ellipsoid holding which holds all the points. We add a 1mm precision margin to the ellipsoid to filter out the noise and outliers. For each point of a strawberry point cloud, we compute the distance from the intersection between the ray passing by the point and the ellipsoid surface. This distance is $\Delta k$ in Eq.~\ref{eq:roundness}. In this formula, a b and c are the ellipsoid parameters, and n the number of points.
\begin{equation}\label{eq:roundness}
    R = 1 - \frac{\sum_{1}^{k}{\Delta k}}{n(abc)^{\frac{1}{3}}}
\end{equation}
In that case 1 means a perfect ellipsoid with lower values indicating sharper and irregular shape. Furthermore as our sensor data is collected from a single view, we compensate for partial views by using only half of the ellipsoid; for the SFM data, both sides are taken into account. The roundness values, due to their averaging characteristics, are not affected by the partial visibility. We also align all the points along the z axis based on their longer axis, which should always correspond to the vertical axis of the berry.

We randomly select 40 strawberries from each dataset presented in Table~\ref{tab:datares} and use all 10 high-quality strawberries from~\cite{boli}. For the simulated berries, we consider two cases with and without the reflectance added. For all of the instances, we manually clean up the point clouds to remove potential discrepancies due to misalignment of colour and depth images and noisy outliers, to only keep the points being part of the strawberry shape/surface.

Due to its averaging characteristics, the roundness metric does not express all the perturbations and spikes of noise, nor small surface deformations. We propose to use the same technique as in~\cite{ukrasjustin} to obtain a more precise description of the shape and its surface.
We compute spherical harmonics coefficients as well as roundness values and ellipsoid raddii and present the results in Sec.~\ref{sec:shapequanti}.

To illustrate the ability of the RGBD cameras to render shape information, we compare them against two other sensing techniques and strategies using normal change rate information. This feature is a good visual indicator to evaluate smoothness of surfaces, or in case of sensing failure, noisy or too flat areas indicating troubles to capture surface and shape information correctly.
The normal change rate $N$ indicates the gradient in the local surface around each point, expressed as 
\begin{equation}\label{eq:normalsurface}
    N = \frac{min(e_{1},e_{2},e_{3})}{\sum_{i=1}^{3}{e_{i}} }
\end{equation}
with $e_{1},e_{2},e_{3}$ being the three eigen values from the covariance matrix of the local neighborhood of each point.

\section{Results}
\subsection{Segmentation results}

Table~\ref{tab:res1} contains results for different variants of the algorithms run on the three datasets representing stereo, ToF and simulated sensors. We compare our proposed algorithm CNN3D to the 3D PointNet++ baseline (PNet) and 2D baseline (SegNet). We also consider additional network configurations combining 3D and colour information resulting in CNN3D\textsubscript{C} and PNet\textsubscript{C} variants respectively. 

The 2D baseline which we use as an ultimate reference is characterised by the exceptional performance in our application thanks to its reliance on context-rich visual features such as colour or texture, enabling superior segmentation results. Whilst using 3D information only for segmentation of soft fruit cannot match the 2D baseline, our proposed CNN3D architecture bridges the performance gap between SegNet and PointNet++ by doubling the AUC metrics for each type of dataset. The particularly poor results for PNet are consistent with our initial findings reported in~\cite{visapp20}. For the simulated dataset, where the quality of 3D data is unaffected by the imperfections of sensing technology, we can see that the performance difference between SegNet and CNN3D disappears almost completely (0.01 of AUC) indicating a good discriminative potential of the shape information for our segmentation task. The results also indicate that the overall quality of 3D information from the ToF sensor is superior to that of the stereo (e.g. 0.08 AUC difference for CNN3D).

Adding reflectance to the simulation bring down the results obtained with all the algorithms, to levels closer to those found with ToF dataset. It shows with the previous shape study that taking into account the high reflectance of these fruits help us understand part of the problem with sensing technologies. 

The importance of colour information for our application, is highlighted in the performance of the extended variants of the presented 3D networks (i.e. CNN3D\textsubscript{C} and PNet\textsubscript{C}). The AUC metric is boosted by more than 0.22 for CNN3D\textsubscript{C} and more than 0.41 for PNet\textsubscript{C} when compared to 3D only networks. CNN3D\textsubscript{C} matches and, especially in the simulated case, slightly exceeds the performance of the SegNet baseline. Whilst the use of the proposed combined 3D and colour architecture does not bring dramatic changes in segmentation performance, it can be exploited in systems which estimate other properties of soft fruit such as their 3D pose and shape quality.

\begin{table*}[ht]
\caption{Performance of the proposed CNN3D architecture compared to 2D and 3D baselines.}\label{tab:res1} 
\centering
\begin{tabular}{|c|c|c|c|c|c|}
  \hline
 Model  & Camera & Acc [\%] & $\kappa$ & mean IoU & AUC\\
 \hline
SegNet                  & \multirow{5}{*}{stereo} &   \textbf{98.55}     &   \textbf{0.87}  &  0.77  & \textbf{0.89}  \\
PNet                    &    &   89.35    &   0.45  &  0.19 & 0.31   \\
CNN3D                   &    &   92.11    &   0.62  &  0.42 & 0.62 \\
PNet\textsubscript{C}   &    &   94.74   &   0.69  &  0.52 & 0.72   \\
CNN3D\textsubscript{C}  &    &   97.45    &   \textbf{0.87}  &  \textbf{0.78} & 0.87 \\

\hline
SegNet           &  \multirow{5}{*}{ToF}           &   \textbf{99.17}      &   0.87  &  0.78 & \textbf{0.92}  \\
PNet             &            &   81.78      &   0.42  &  0.19  & 0.29  \\
CNN3D            &            &   95.66      &   0.72  &  0.54  & 0.70  \\
PNet\textsubscript{C}       &            &   92.36      &   0.71  &  0.58  & 0.78   \\
CNN3D\textsubscript{C}      &            &   98.68     &   \textbf{0.91}  &  \textbf{0.85} & \textbf{0.92} \\

 \hline
 SegNet   & \multirow{5}{*}{simu $- $ reflectance}       &   99.56      &   0.96    &  0.92  & 0.98\\
 PNet     &        &   96.61      &    0.63   &   0.39 & 0.57\\
 CNN3D    &        &   99.55      &    0.95   &   0.92 & 0.96\\
 PNet\textsubscript{C}  &     &   97.50      &   0.74    &  0.60 & 0.82\\
 CNN3D\textsubscript{C} &     &   \textbf{99.72}     &  \textbf{0.98}    &  \textbf{0.95}  & \textbf{0.99}\\
 \hline
SegNet   & \multirow{5}{*}{simu $w/$ reflectance }       &   \textbf{99.56}      &   \textbf{0.96}    &  0.92  & \textbf{0.98}\\
 PNet     &        &   95.77      &    0.43  &   0.11 & 0.19\\
 CNN3D    &        &  99.11 & 0.92 & 0.86 & 0.93 \\
 PNet\textsubscript{C}  &     &   97.45      &   0.60   &  0.32 & 0.48\\
  CNN3D\textsubscript{C} &     &   99.27     &  0.94    &  \textbf{0.93}  & 0.98\\
 \hline
\end{tabular}
\end{table*}

The proposed CNN3D architecture, due to its nature, is computationally efficient with inference times ($\sim$\SI{22}{\milli\second} per point cloud) comparable to those of SegNet ($\sim$\SI{20}{\milli\second} per image). This is an order of magnitude less than the processing times required by PointNet++ ($\sim$\SI{200}{\milli\second}). The additional step of normal computation can be implemented efficiently using integral images resulting in processing times of $<$\SI{10}{\milli\second}, depending on the complexity of the scene. The reported times are for the parallelised variants of the networks running on an NVidia GPU GTX 1080 Ti.

\begin{figure*}[hb!]
    \centering
    \begin{tabular}{cccc}
        \includegraphics[width=0.2\columnwidth]{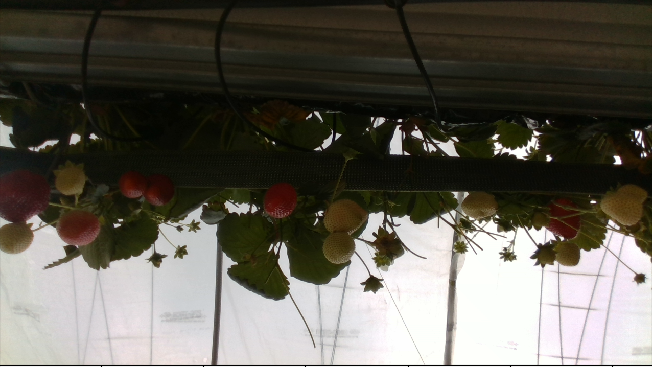}&
        \includegraphics[width=0.2\columnwidth]{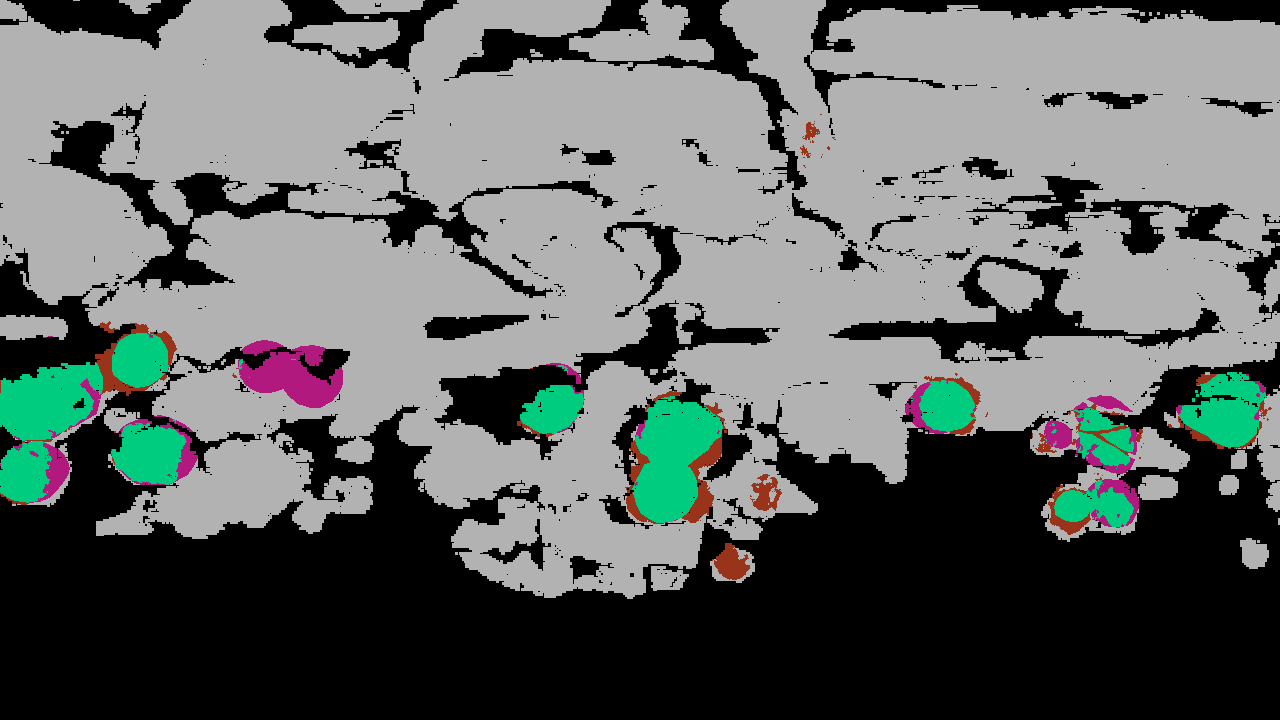}&
        \includegraphics[width=0.2\columnwidth]{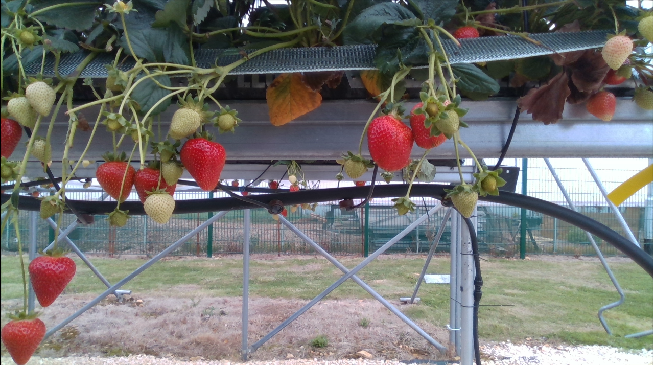}&
        \includegraphics[width=0.2\columnwidth]{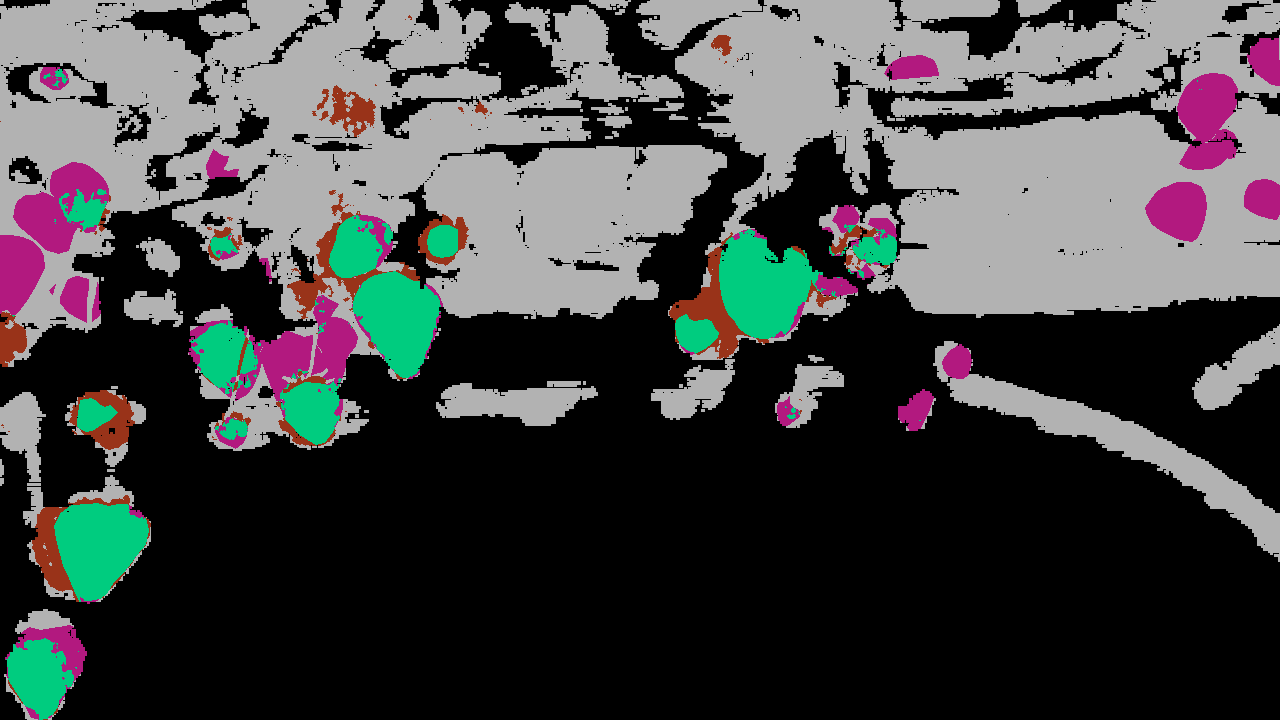}\\

        \includegraphics[width=0.2\columnwidth]{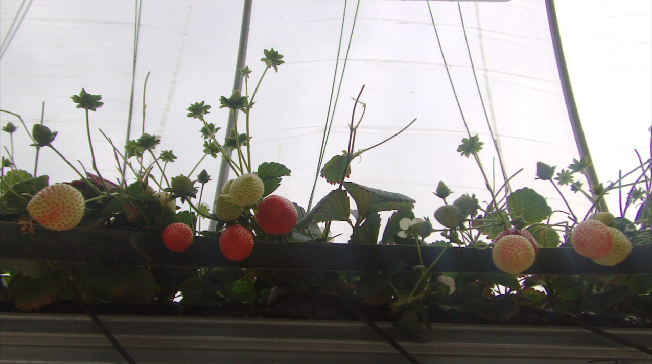}&
        \includegraphics[width=0.2\columnwidth]{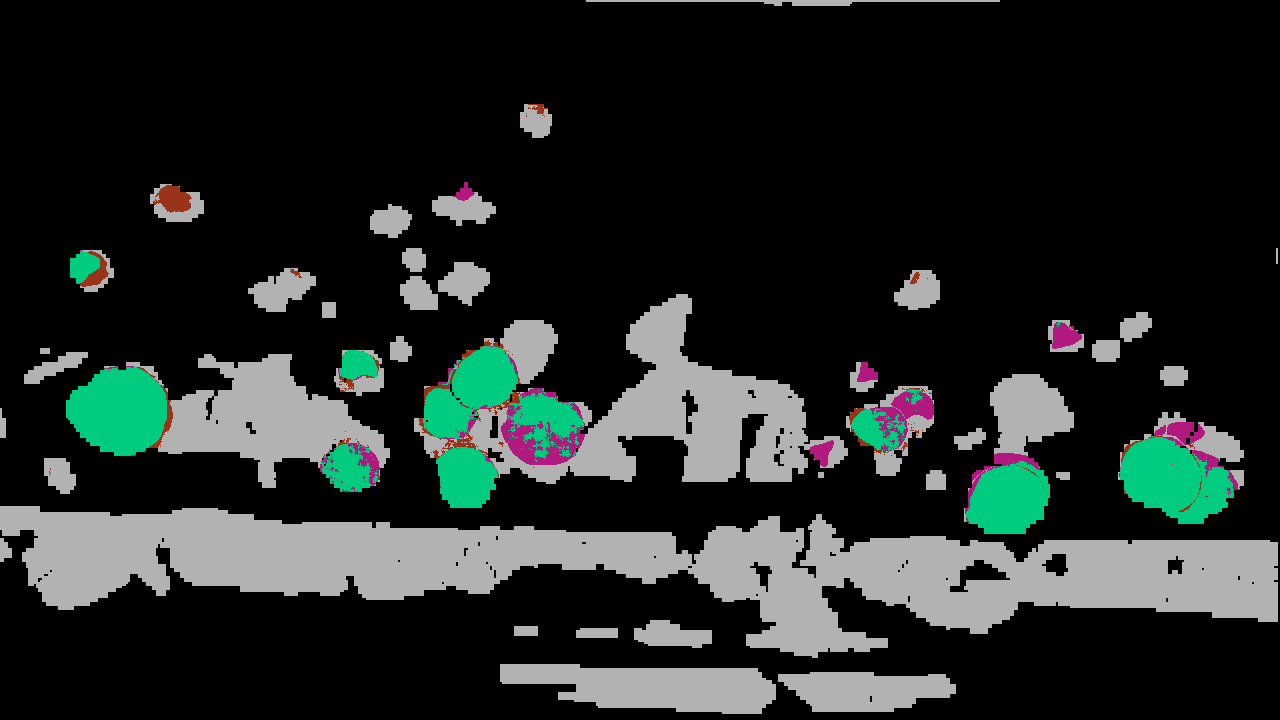}&
        \includegraphics[width=0.2\columnwidth]{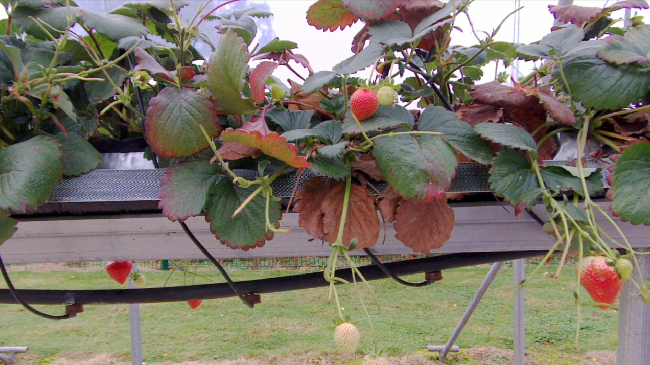}&
        \includegraphics[width=0.2\columnwidth]{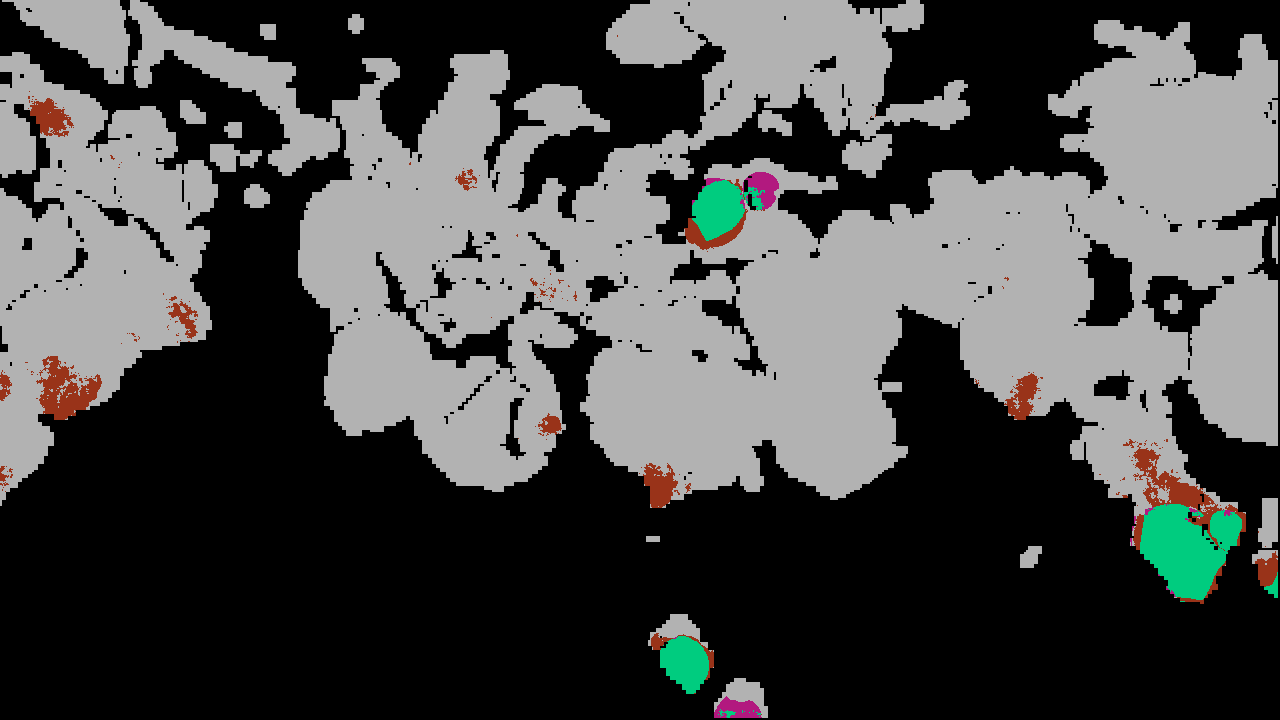}\\
        
        \includegraphics[width=0.2\columnwidth]{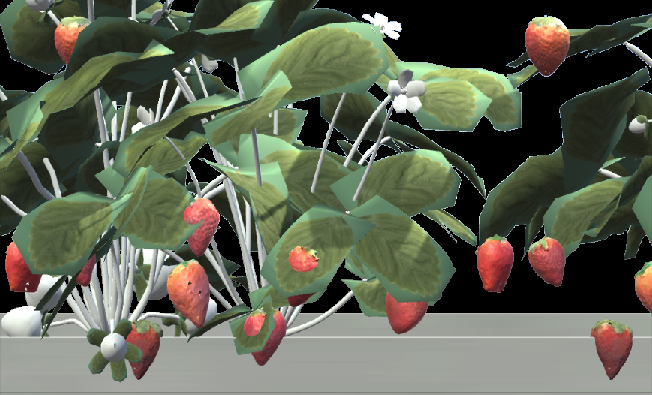}&
         \includegraphics[trim=0cm 12.2cm 0cm 0cm,clip,width=0.2\columnwidth]{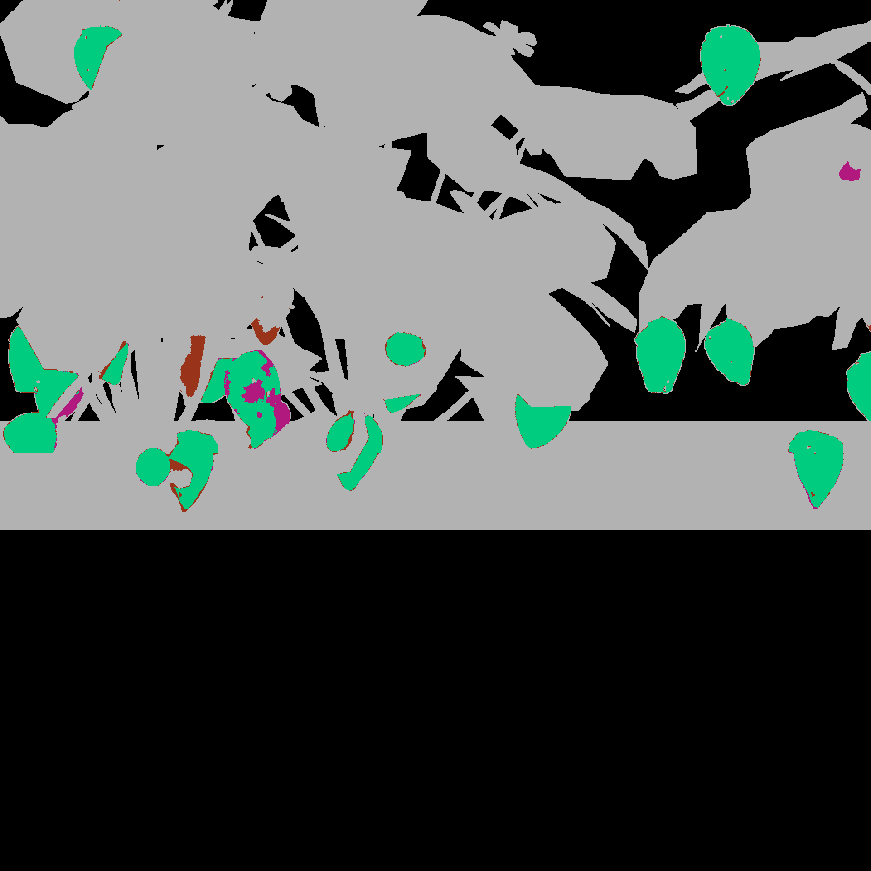}&
        \includegraphics[width=0.2\columnwidth]{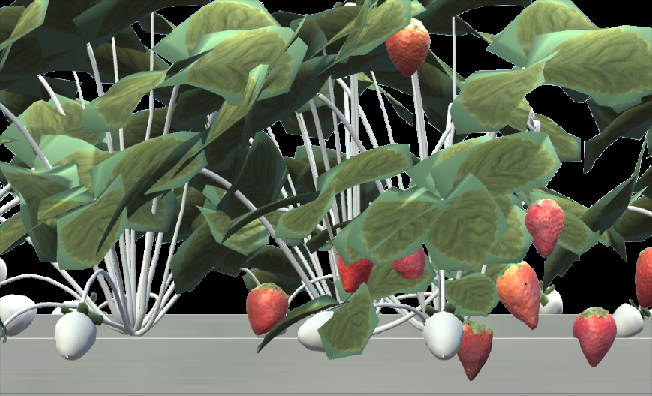}&
        \includegraphics[trim=0cm 12.2cm 0cm 0cm,clip,width=0.2\columnwidth]{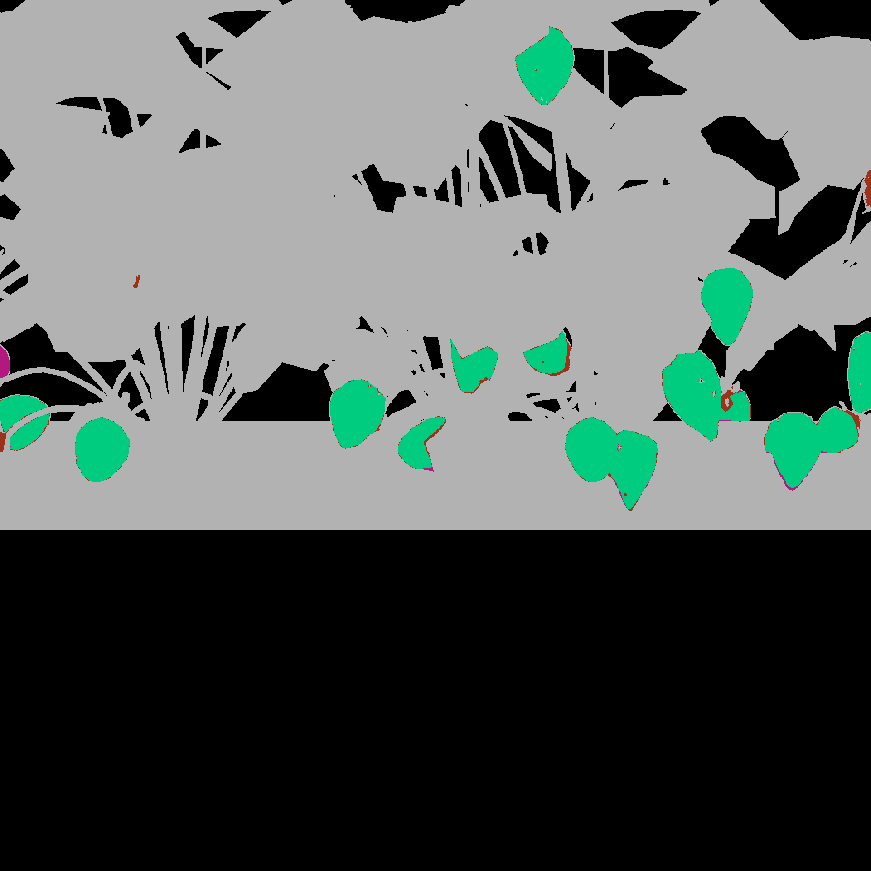}\\
    \end{tabular}
    \caption{The segmentation results for CNN3D trained on data from the stereo (top row), ToF (middle row) and simulated data (bottom row). The colours correspond to TP (green), FP (orange), FN (purple), TN (gray) and no data (black). Left column is an underneath view while the right side is a side view (except for the simulation, where both are side view).}
    \label{fig:quali1}
\end{figure*}

Fig.~\ref{fig:quali1} presents the output examples of the proposed CNN3D method applied to the three datasets. The examples illustrate the overall good quality of the predictions and suitability of 3D information for the segmentation task. The problematic regions match closely the observations made in Sec.~\ref{sec:sensors3d}. The relatively high number of False Negatives for the stereo data is corresponding to flat surfaces whilst the higher noise profile for the ToF sensor results in increased False Positives. In the simulated data, the areas most susceptible to wrong segmentation are those heavily occluded and those corresponding to fruits with very particular pose and shape not present in the training set.

\begin{figure}[ht!]
    \centering
    \begin{tabular}{c|c|c}
        \includegraphics[width=0.3\columnwidth,height=0.3\columnwidth]{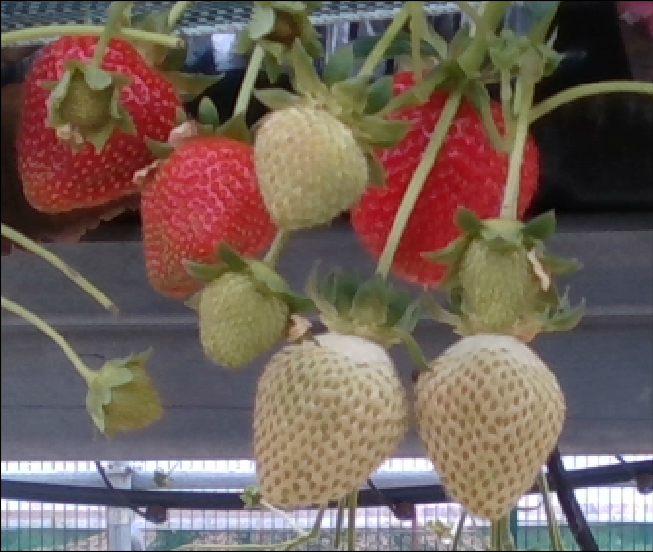}&
        \includegraphics[width=0.3\columnwidth,height=0.3\columnwidth]{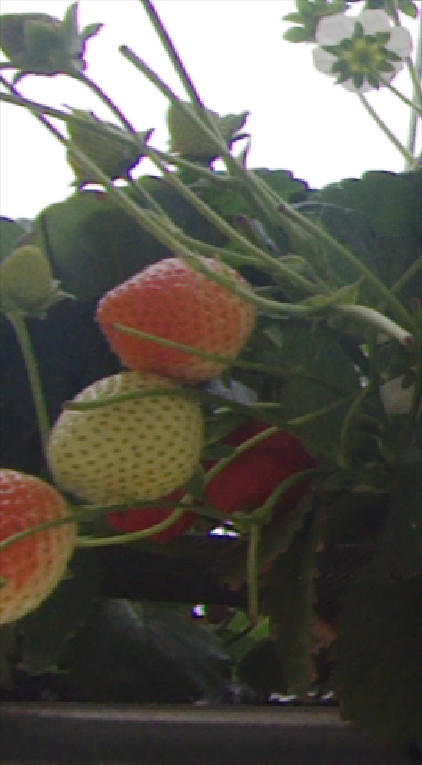}&
        \includegraphics[width=0.3\columnwidth,height=0.3\columnwidth]{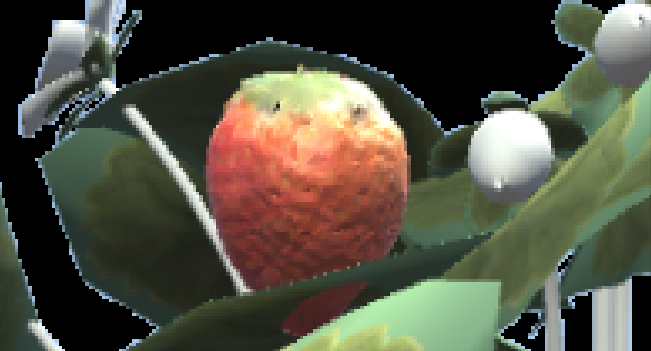}\\
        \includegraphics[width=0.3\columnwidth,height=0.3\columnwidth]{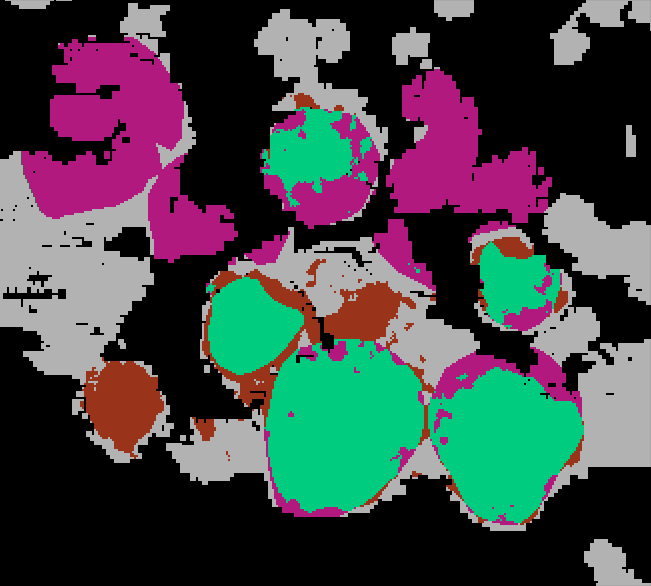}&
        \includegraphics[width=0.3\columnwidth,height=0.3\columnwidth]{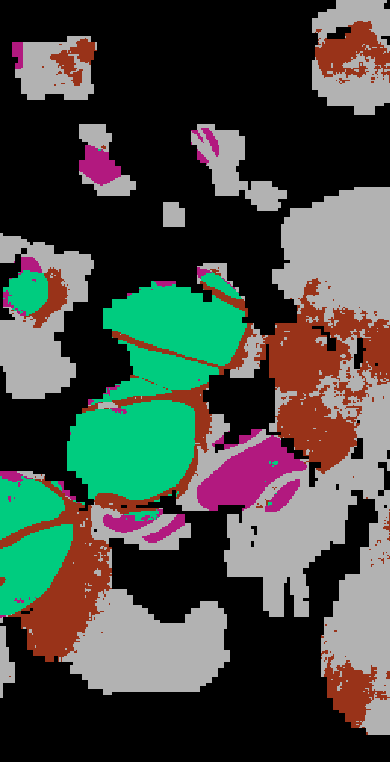}&
        \includegraphics[width=0.3\columnwidth,height=0.3\columnwidth]{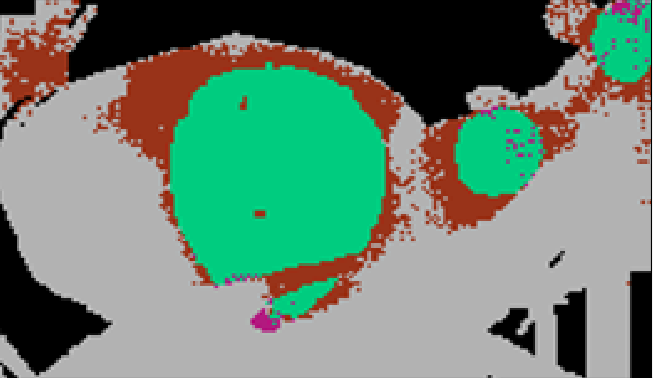}\\
    \end{tabular}
    \caption{The fine detail of segmentation results for particularly difficult cases for CNN3D trained on data from the stereo (left), ToF (centre) and simulated data (right). The colours correspond to TP (green), FP (orange), FN (purple), TN (gray) and no data (black).}
    \label{fig:quali2}
\end{figure}

We present zoomed-in examples in Fig.~\ref{fig:quali2}, for each sensors. The same characteristics of errors appear, with stereo data lacking shape and increasing False Negatives, ToF creating noisier areas increasing False Positives, and the noise added to the simulated data, blurring areas around the strawberries creating less precise object contours.

\subsection{CNN3D ablation} 
We study the model improvement we added, with the study of the conv transpose effect and normal features added as an input. We train CNN3D with a conv and transpose convolution (in decoder part) variation on simulated data with and without reflectance, with points as only inputs and points and surface normals combined. We present the results in Tab.~\ref{tab:abla}.
\begin{table*}[ht!]
    \centering
    \begin{tabular}{|c|c|c|c|c|c|}
        \hline
         model  & Camera&  Acc & $\kappa$ & mean IoU & AUC   \\
        \hline
         pts$_{c}$ & \multirow{2}{*}{stereo}& 90.78\% & 0.61 & 0.40 & \textbf{0.63}\\
         pts$_{t}$ & & 91.00\% & 0.60 & 0.36 & 0.61\\
         pts$+$norm\textsubscript{c}      & & \textbf{92.11}\% & 0.62 & 0.36 & 0.61\\
         pts$+$norm\textsubscript{t} & & 91.39\% & \textbf{0.62} & \textbf{0.42} & 0.62\\
        \hline
          pts$_{c}$ &\multirow{2}{*}{ToF} & 92.25\% & 0.57 & 0.32 & 0.48\\
         pts$_{t}$ & & 93.67\% & 0.63 & 0.39 & 0.58\\
         pts$+$norm\textsubscript{c} && 95.07\% & 0.70 & 0.53 & 0.70\\
         pts$+$norm\textsubscript{t}& &\textbf{95.66\%} & \textbf{0.72} & \textbf{0.54} & \textbf{0.70} \\
        
         \hline
         pts\textsubscript{c}                & \multirow{4}{*}{simu $-$ reflectance}& 99.09\% &  0.91 & 0.84 & 0.92\\
         pts\textsubscript{t}           & & 98.54\% & 0.86 & 0.76 & 0.88\\
         pts$+$norm\textsubscript{c}      & & 99.21\% & 0.92 & 0.86 & 0.93\\
         pts$+$norm\textsubscript{t} & & \textbf{99.55\%} & \textbf{0.95} &  \textbf{0.92} & \textbf{0.96}\\
        \hline
         pts\textsubscript{c}                &\multirow{4}{*}{simu $w/$ reflectance}& 98.78\% & 0.88 & 0.79 & 0.90\\
         pts\textsubscript{t}           & & 98.71\% & 0.87 &  0.78 & 0.89\\
         pts$+$norm\textsubscript{c}      & & \textbf{99.16\%} & 0.92 & 0.86 & 0.93 \\
         pts$+$norm\textsubscript{t} & & 99.11\% & \textbf{0.92} &\textbf{ 0.86} & \textbf{0.93}\\
        \hline
         
    \end{tabular}
    \caption{Ablation study of the effect inputs : pts (points coordinates) and norm (normals), and conv transpose (\textsubscript{t}) instead of convolutions (\textsubscript{c})}
    \label{tab:abla}
\end{table*}

Using normals as extra input feature, consistently improve the results across simulation variations.
conv transpose on the other hand struggle with noisier inputs with reflectance added. This is logical with the nature of conv transpose, as with reflectance, local information becomes more unreliable and spread feature maps less an advantage. However with stereo and ToF dataset, conv transpose proves to be more suited and improving the results compare to standard convolutions.

\subsection{Quantitative shape evaluation}\label{sec:shapequanti}
We study the shape of the strawberries across our four captured datasets (simu $-$ reflectance,simu $w/$ reflectance,ToF,stereo), and for comparison the high quality data captured in~\cite{boli}. For this purpose we propose studying roundness of their shape, their bounding ellipsoid characteristics, as well as their spherical harmonics characteristics.

We report the results in Table~\ref{tab:shaperes}, which denotes the roundness values (average with std) and average ellipsoid radii/parameters for each data set. The ellipsoid parameters are normalised for direct comparison between the different datasets. The ToF and simulation (without reflectance) produce similar roundness values and ellipsoid shapes, which confirms overall better shape quality obtained with ToF sensing. On the other hand the stereo sensor produces very flat surfaces, which are also visible in the simulated results including the reflectance (due to surface saturation). 

\begin{table}[ht!]
        \caption{The average shape characteristics for each data set.}\label{tab:shaperes} 
        \begin{tabular}{|c|c|c|}
              \hline
              data & roundness & ellipsoid radii \\
             \hline
              stereo & $0.23\pm0.11$ & [0.15, 0.62, 0.77]   \\
              ToF & $0.51\pm0.10$ & [0.38, 0.56, 0.73] \\
              simu $-$ reflectance  & $0.52\pm0.07$ & [0.42, 0.55, 0.72]\\
              simu $w/$ reflectance & $0.29\pm0.12$ & [0.29, 0.49, 0.82]\\
              \cite{boli} & $0.80\pm0.02$ & [0.47, 0.53, 0.70]\\
         \hline
        \end{tabular}
\end{table}

\begin{figure*}[ht!]
\centering
        \begin{tabular}{ccc}
            \includegraphics[trim=1.2cm 0 1.5cm 0, clip,width=0.32\columnwidth]{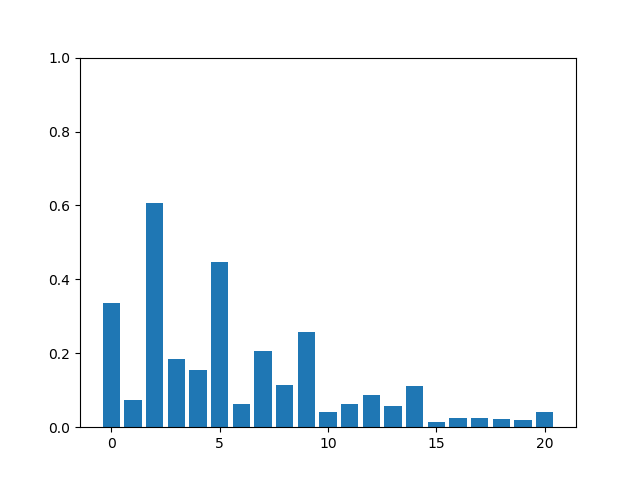}&
            \includegraphics[trim=1.2cm 0 1.5cm 0, clip,width=0.32\columnwidth]{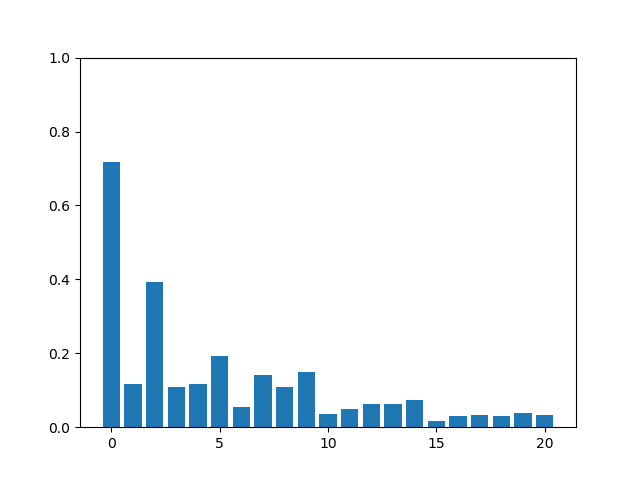}&
            \includegraphics[trim=1.2cm 0 1.5cm 0, clip,width=0.32\columnwidth]{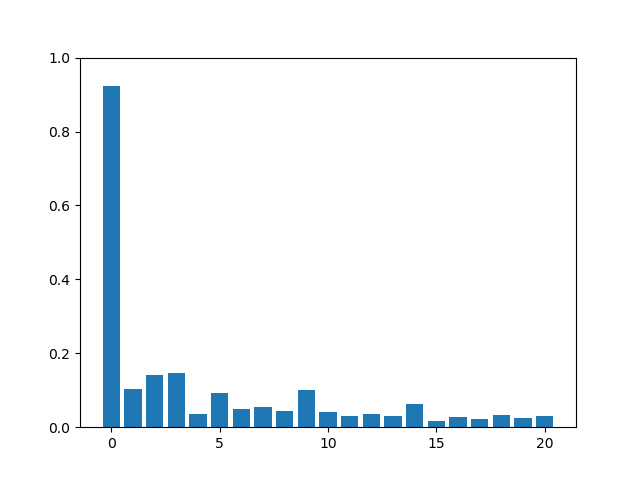}\\
            Stereo & ToF  & Simu$-$reflectance \\
        \end{tabular}
        \begin{tabular}{cc}
            \includegraphics[trim=1.2cm 0 1.5cm 0, clip,width=0.32\columnwidth]{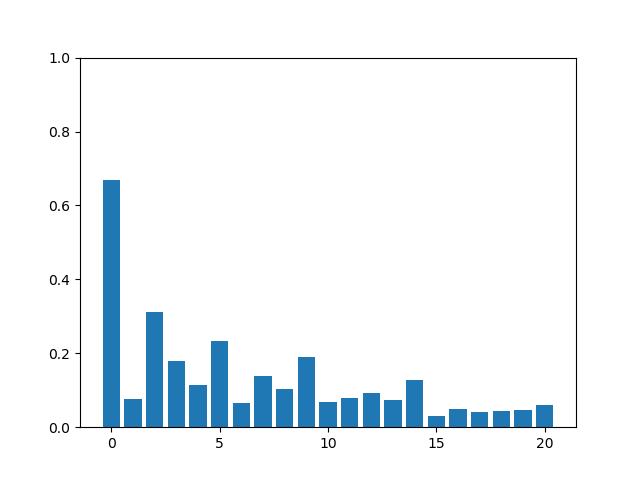}&
            \includegraphics[trim=1.2cm 0 1.5cm 0, clip,width=0.32\columnwidth]{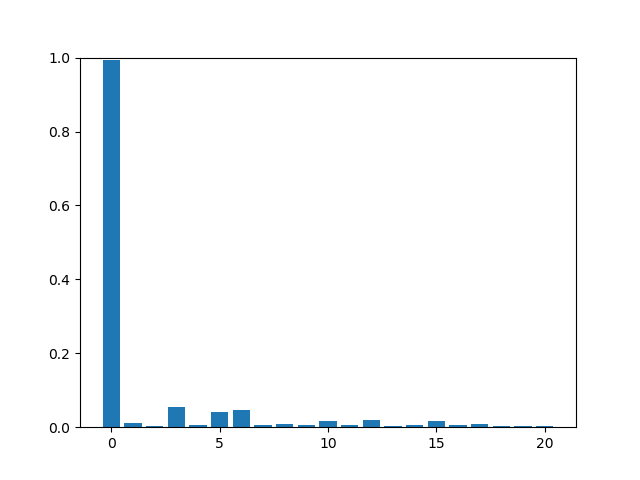}\\
            Simu$w/$reflectance & \cite{boli}\\
        \end{tabular}
    \caption{Spectral profiles of strawberry point clouds for each dataset.} \label{fig:sph}
\end{figure*}

 We compute the backward transformation of the spherical harmonics for all the berries and average the coefficients obtained to create an overall spectral profile for each set (see Fig.~\ref{fig:sph}). With Gaussian noise only, the simulation spectrum is relatively close to that of the SFM data. We see similar spectra for the simulation and ToF with higher perturbations for ToF. Spectral profile for the stereo data is characterised by a far greater noise level and deformed shape. Once reflectance is added, the simulation profile becomes closer to the ToF profile indicating higher noise levels.

\subsection{Qualitative shape evaluation}\label{sec:qualishape}
The final shape evaluation we offer is qualitative and relies on the normal change rate information. For this purpose we compare crops from two other datasets with data capture using stereo and ToF sensors on strawberries. We also look at the influence of external light by capturing data indoor with both sensors.

\begin{figure}[!ht]
\centering
\begin{tabular}{c|c}
    \includegraphics[width=0.49\columnwidth,height=0.4\columnwidth]{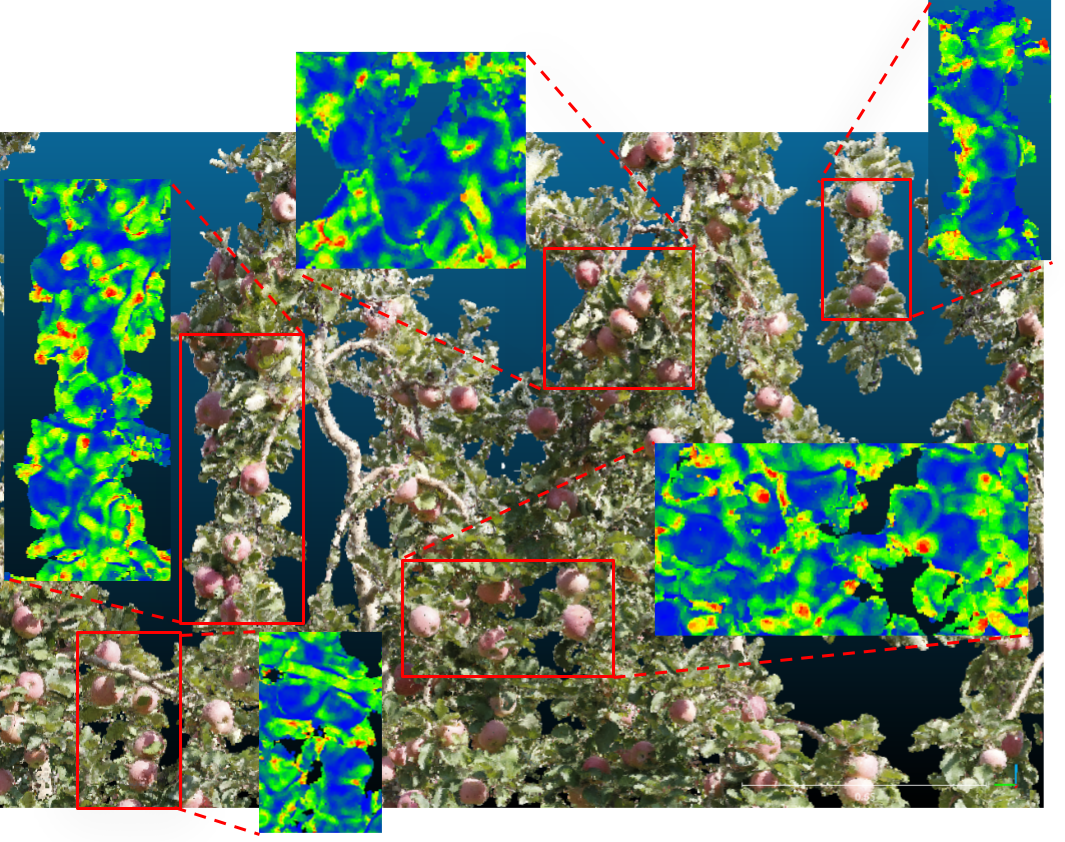}&
    \includegraphics[width=0.49\columnwidth,height=0.4\columnwidth]{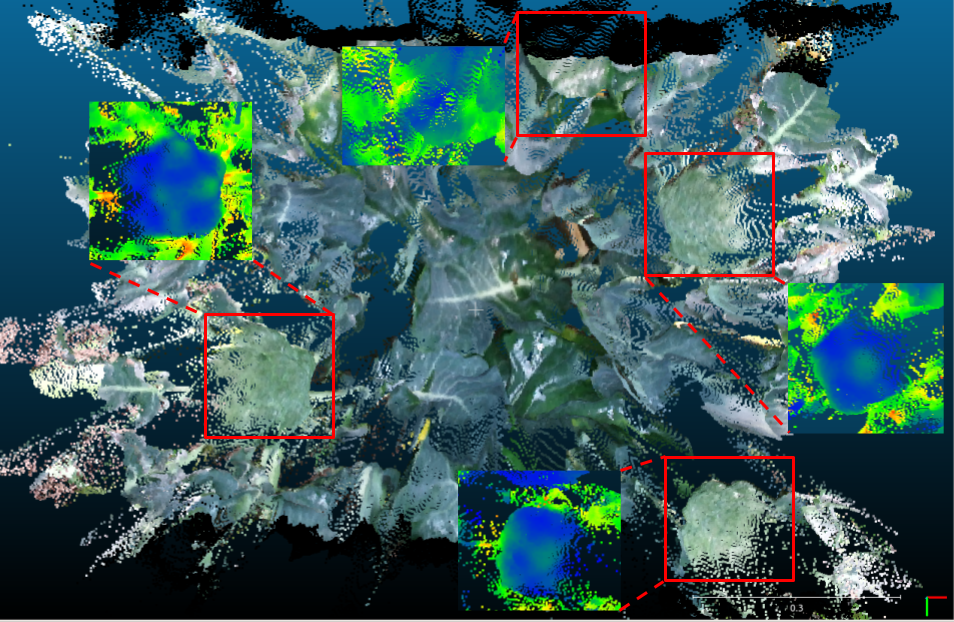}\\
    A&B\\
    \includegraphics[width=0.49\columnwidth,height=0.4\columnwidth]{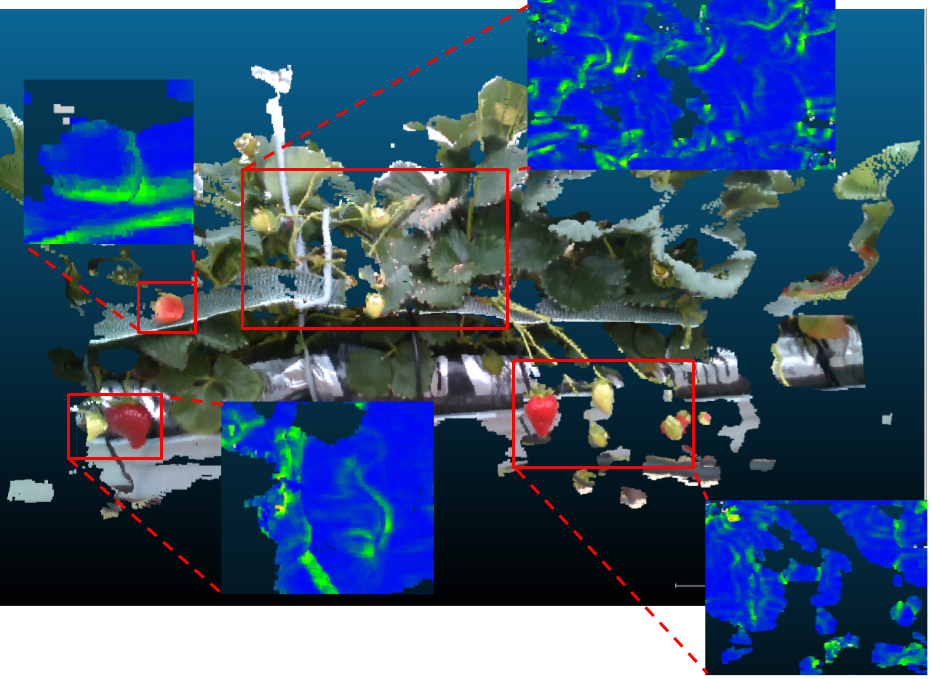}&
    \includegraphics[width=0.49\columnwidth,height=0.4\columnwidth]{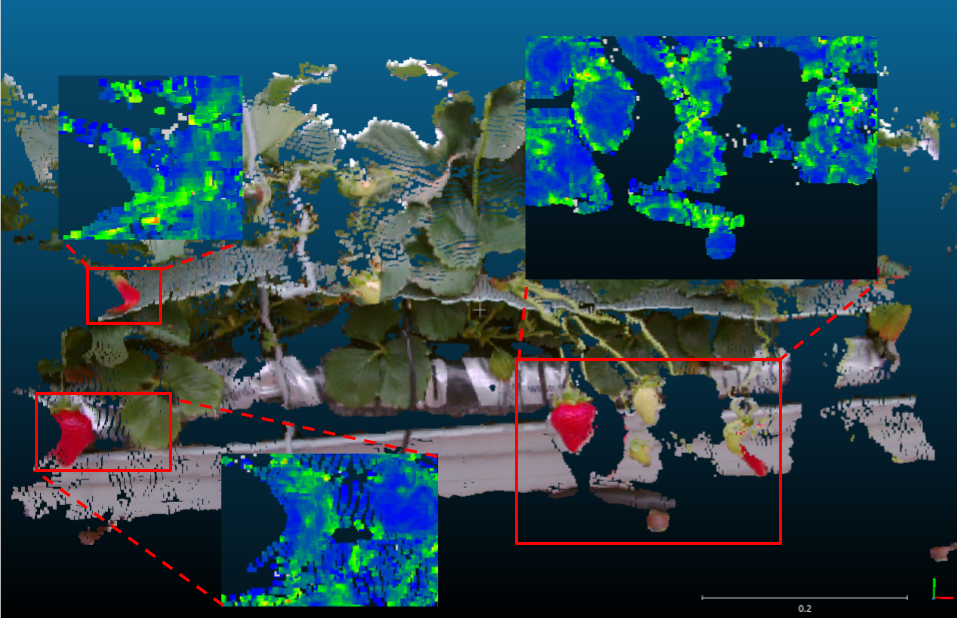}\\
     C&D\\
\end{tabular}
\caption{Comparison between different fruits/plants and sensing techniques, with their pointcloud and RGB image, and the normal change rate for shape and surface analysis: (A) Fuji apple dataset, (B) broccoli dataset, (C) strawberry stereo, (D) straberry ToF.} \label{fig:fruitcomparison}
\end{figure}

We present in Fig.~\ref{fig:fruitcomparison} a point cloud shape comparison from 4 different datasets captured using different methods. The first example is from the Fuji apple dataset~\cite{GENEMOLA2020105591}, where a point cloud representing an apple orchard was created using RGB images and the structure-from-motion (SFM) algorithm. The second example is from a broccoli dataset \cite{kusuman}, captured using the Kinect sensor, and an enclosing box set-up reducing the influence of the sun on the crop. The two last examples are respectively from the Intel RealSense stereo camera and the Pico Zense ToF camera, captured at the same time with sensors stacked on top of each other to offer the same capturing conditions. Both the Fuji apple and broccoli datasets offer a different perspective on the sensing of objects in agricultural context. The former does not rely on any 3D camera to get spatial and shape information but rather large amount of RGB images, while the later minimises sun exposure through a pseudo-controlled environment to enhance the quality of the data collected. These choices improve largely the quality of the sensed shape of the objects of interest, but also come with additional limitations, which are not easily applicable in soft fruit scenarios. Firstly using SFM requires large amount of images and processing power to register them into a point cloud, which limits the use of SFM in real time applications. Concerning the broccoli dataset, the solution relies on the fact that broccoli plants grow on the ground where using box enclosures is fairly straightforward. This is much more difficult for strawberries, which are vertically hanging from table-tops where any enclosure would need block the sun off from all sides resulting in significant size of the enclosure.

The other factors which should be considered are both the size and reflectiveness of the strawberry compared to crops such as apples or broccoli. These are directly observable in the two examples presented in Fig.~\ref{fig:fruitcomparison}. The data coming out of the stereo camera is lacking details, becoming very flat with a clear lack of shape features and with the fruit often indistinguishable from the background. The ToF data better represents the shape, but reflectiveness becomes a problem resulting in noisier information with additional noise on the surface and a slightly deformed shape, while having better boundary definition between the objects leading to a better chance for distinguishing the fruits from the background.

\begin{figure}[!ht]
\centering
\begin{tabular}{c|c|c|c}
    \multicolumn{4}{c}{\includegraphics[width=0.8\columnwidth,height=0.3\columnwidth]{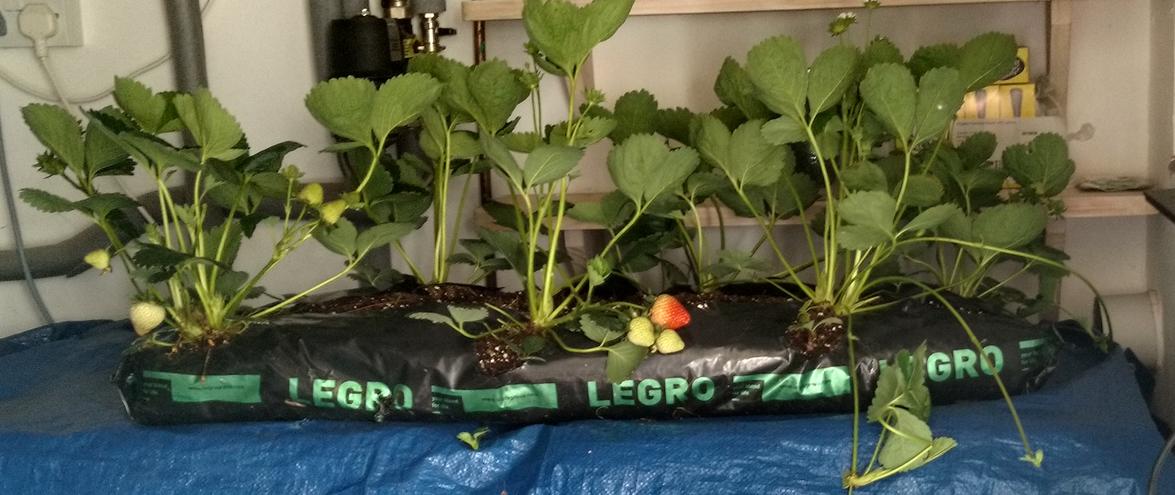}}\\
    \includegraphics[width=0.21\columnwidth,height=0.2\columnwidth]{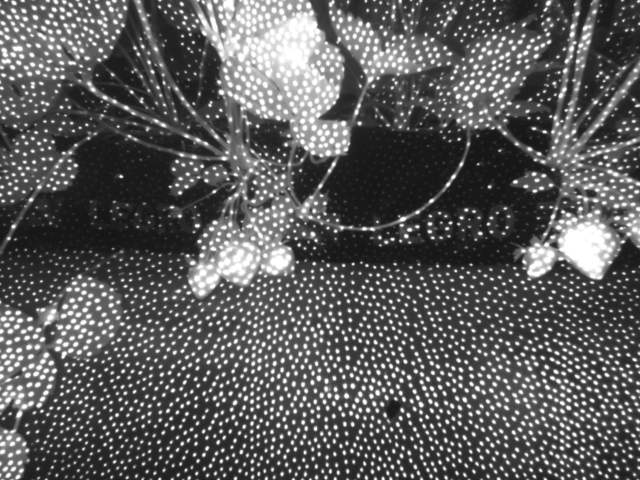}&
    \includegraphics[width=0.21\columnwidth,height=0.2\columnwidth]{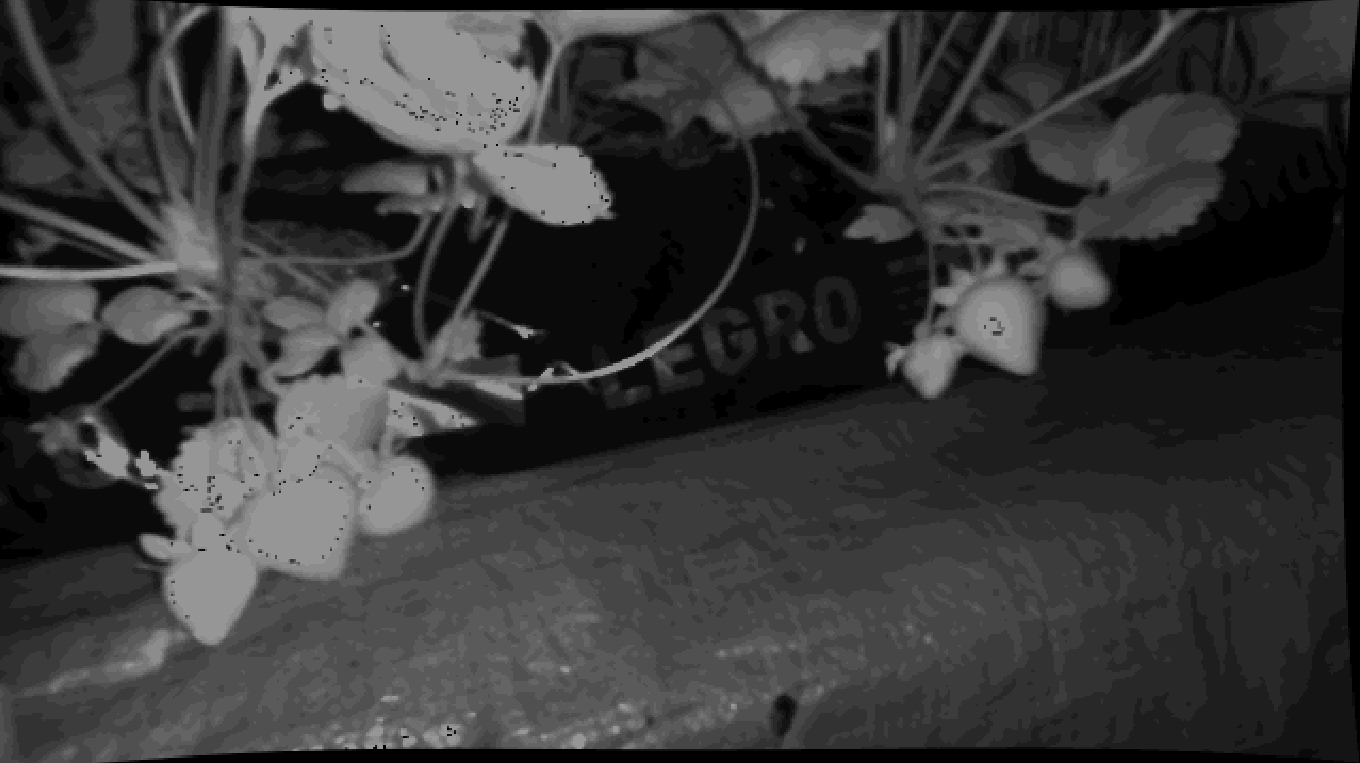}&
    \includegraphics[width=0.21\columnwidth,height=0.2\columnwidth]{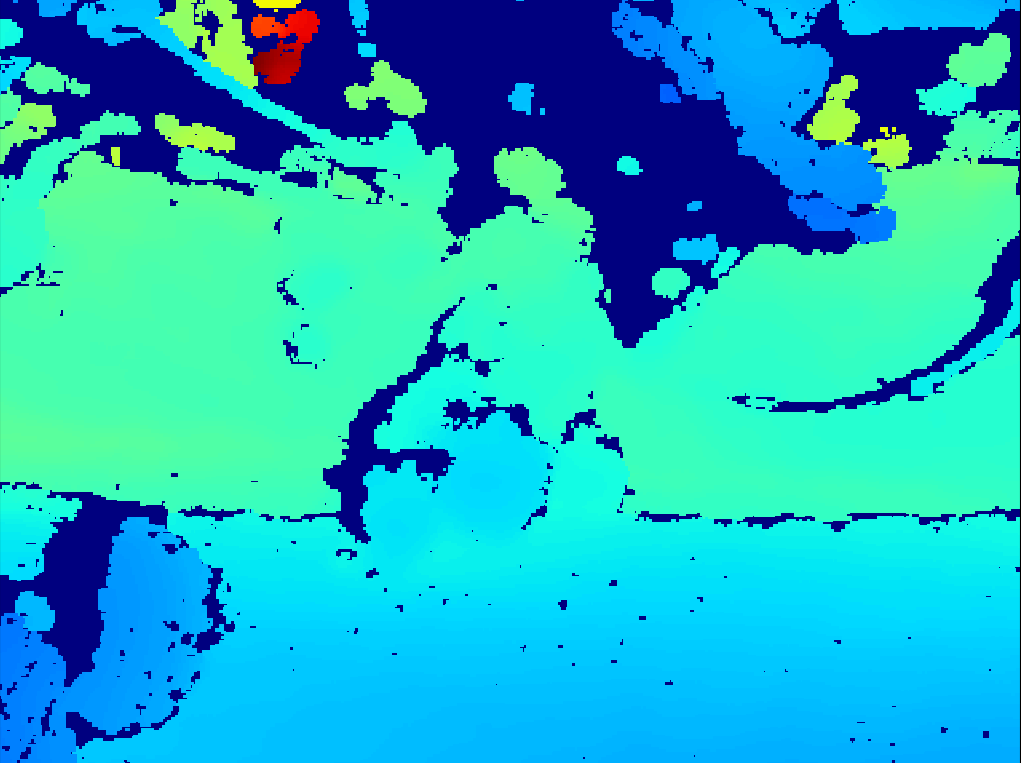}&
    \includegraphics[width=0.21\columnwidth,height=0.2\columnwidth]{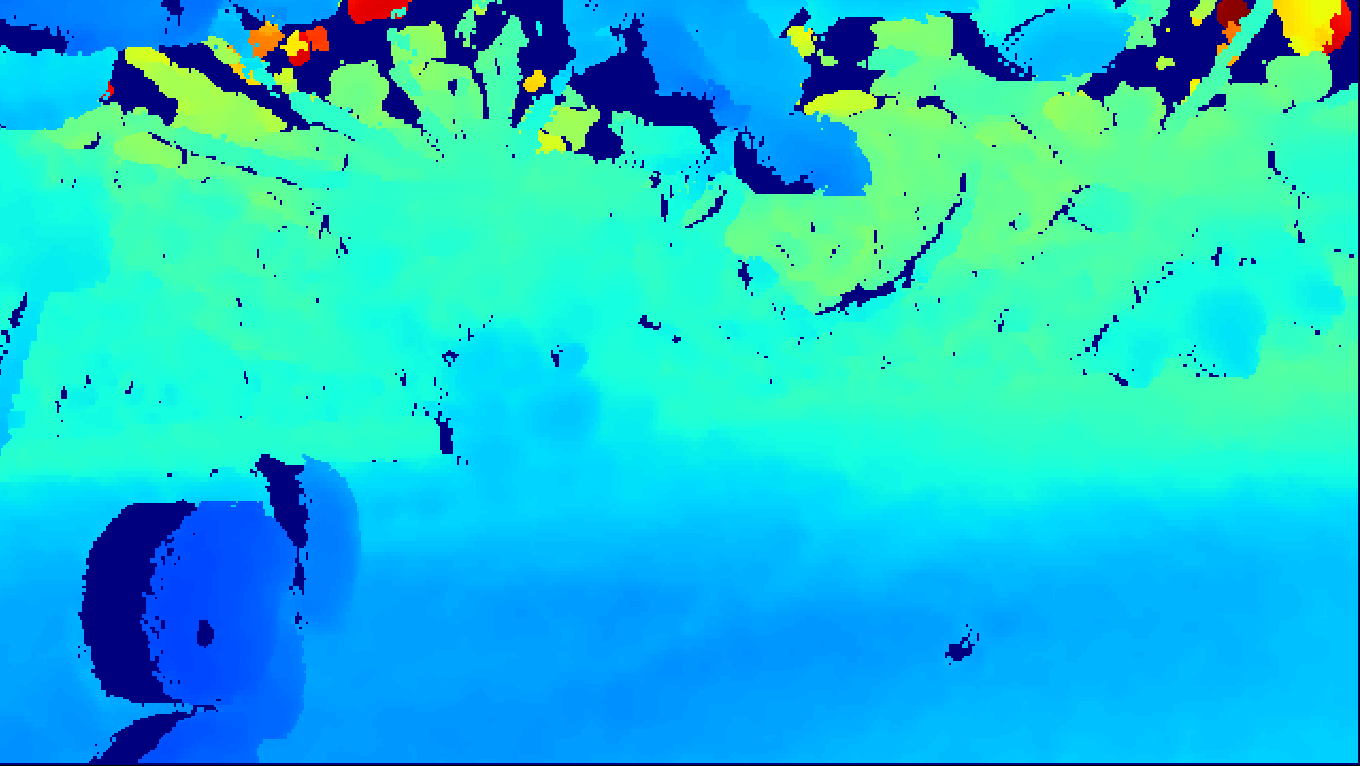}\\
    stereo & ToF & stereo & ToF\\
\end{tabular}
\caption{A side view of the strawberry plants in controlled environment (1st row) as seen through the stereo and ToF sensors including the infra-red (bottom left) and depth (bottom right) images.} \label{fig:indoordata}
\end{figure}
\begin{figure}
\centering
\begin{tabular}{c|c}
    \includegraphics[trim=0cm 4cm 0cm 0cm, clip,width=0.35\columnwidth,height=0.35\columnwidth]{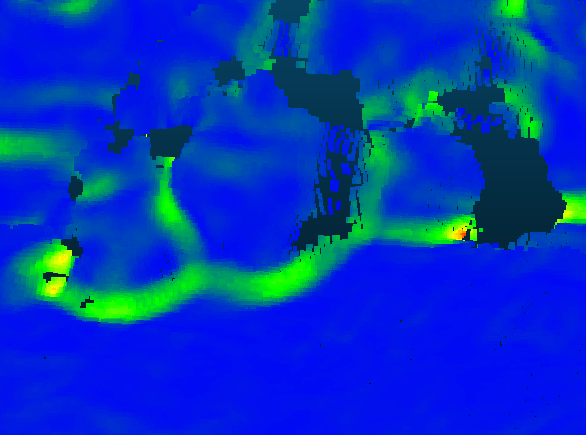}&
    \includegraphics[trim=0cm 2cm 0cm 0cm, clip,width=0.35\columnwidth,height=0.35\columnwidth]{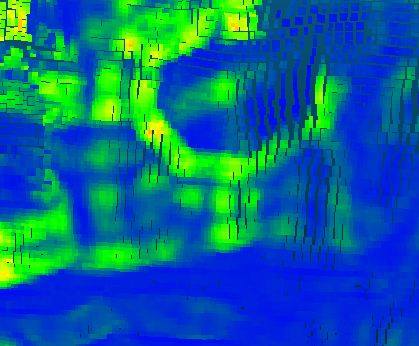}\\
    stereo & ToF\\
\end{tabular}
\caption{Normal change rate of a strawberry cluster showing detail from Fig.~\ref{fig:indoordata} for the stereo (left) and ToF (right) sensors.} \label{fig:detailsindoor}
\end{figure}

To assess the influence of the changing light conditions, we have also captured a snapshot of the same strawberry plants in an indoor, controlled environment where the amount of sunlight is reduced to a minimum using both cameras (see Fig.~\ref{fig:indoordata}). In this setup, the characteristic projected pattern is clearly visible in the infra-red image from the stereo camera. The majority of large surfaces are covered uniformly by the pattern but the reflective and small surfaces of the plant create very high distortion and blend the projected dots resulting in greatly reduced accuracy of depth estimation. The resulting irregularities in depth estimation are of similar scale and shape as the object of interest (i.e. strawberry fruit) and therefore might negatively impact the discriminative capabilities of the object detector. The major challenge for the ToF sensor are reflective surfaces such as the plastic tarpaulins or growing bags and plant parts which saturate and blend together affecting the quality of the reconstructed depth.

We show the normal change rate of the point cloud in Fig.~\ref{fig:detailsindoor}, which again demonstrates the same characteristics as outdoor data. Even without the influence of the sun, the size and reflectance of the fruits seems to be impacting the shape information in the depth images created by the cameras.

Both examples indicate potential challenges in deploying 3D vision systems for detection of small objects in agriculture. These are not only limited to the external fluctuations of infra-red light from the sun but also include reflective objects such as strawberry fruit which are particularly sensitive to infra-red light. The projected pattern of the stereo camera is particularly prone to side effects and poor results in depth estimation. For both sensors, the small scale surfaces are difficult since these do not offer enough features and variation for an accurate depth reconstruction. Getting the cameras closer to the objects could potentially help, but the current technology does not allow for measuring at distances less than 20-30 cm.

\section{Conclusion}

Capturing and utilising 3D information for object detection, segmentation or classification is a challenging task especially in the agricultural context presented in this paper. Our study evaluated two 3D sensing technologies, offered a study of the shape produced by them, proposed an efficient network segmentation architecture and  compared its performance to 3D and 2D variants of state-of-the-art neural networks trained on the data collected from a real strawberry farm and its realistic representation in simulation. These results show encouraging performance but also allow us to highlight the limitations of current technologies and algorithms. Time-of-Flight technology, despite its superior quality of point clouds and shape information, struggles with reflective surfaces resulting in a large number of false segmentations, while stereo technology, lacking detail in acquired depth, fails to detect numerous fruits. On the other hand, the proposed CNN3D network narrows the gap between traditional 2D image-based convolutional neural networks and 3D architectures for the segmentation task and demonstrates comparable performance in simulated scenarios indicating future promise of employing 3D information for such applications. Using 3D information for detection and segmentation is becoming a very important approach to consider in the agricultural domain. And sensing good shape quality is an important factor to consider when deploying robots into the field. It could alleviate problems such as  occlusions, clusters and very close objects, faced while considering only colour images for detection. Implementing reliable transfer learning between realistic simulation and real world sensor data, is the ultimate goal and the subject of our future work. First understanding completely the challenges faced by the sensing technologies and adapting the simulation to solve them, before trying to bridge the gap between simulation and real world. This work should encourage researchers and companies to develop more accurate and robust 3D sensing technologies benefiting future applications in agriculture.

\bibliography{cas-refs}

\end{document}